\documentclass[10pt,twocolumn,letterpaper]{article}

\usepackage{cvpr}              %

\usepackage[export]{adjustbox}
\usepackage{graphicx}
\usepackage{caption}
\usepackage{subcaption}
\usepackage{printlen}
\usepackage{float}
\usepackage{wrapfig}
\usepackage{multirow}
\usepackage{booktabs}
\usepackage{pifont}
\usepackage{makecell}
\usepackage{nicefrac}
\usepackage{url}
\usepackage{yhmath}
\usepackage{stmaryrd}
\usepackage{verbatim}
\usepackage{amsmath}
\usepackage{algorithm}
\usepackage{algpseudocode}
\usepackage{amsfonts}
\usepackage{microtype}
\usepackage{pgf}
\usepackage{pgfplots}
\usepackage{tabularx}
\usepackage{ifthen}
\usepackage{soul}
\usepackage{tikz}
\usepackage{xspace}
\usepackage[utf8]{inputenc}
\usepackage[T1]{fontenc}
\usetikzlibrary{angles,quotes,3d,math,arrows.meta,calc,positioning,fit,backgrounds,decorations.pathreplacing,calligraphy,shapes,shapes.multipart}
\usepackage[table]{xcolor}
\usepackage[accsupp]{axessibility}
\usepackage{cuted}

\renewcommand{\paragraph}[1]{\vspace{.5em}\noindent\textbf{#1.}}
\newcommand\rurl[1]{%
  \href{https://#1}{\nolinkurl{#1}}%
}

\newcommand{\supi}[1][i]{\ensuremath{^{\smash{(#1)}}}} %

\newcommand{\cmark}{\ding{51}}
\newcommand{\xmark}{\ding{55}}

\newcommand{\modelname}{Myriad\xspace}
\newcommand{\ourdataname}{OWM\xspace}

\newcommand{\billiardsamples}[1]{#1}

\definecolor{ourgreen}{RGB}{46, 204, 113}
\definecolor{ourgreenborder}{RGB}{39, 174, 96}
\definecolor{ourblue}{RGB}{52, 152, 219}
\definecolor{ourblueborder}{RGB}{41, 128, 185}
\definecolor{ourorange}{RGB}{230, 126, 34}
\definecolor{ourorangeborder}{RGB}{211, 84, 0}
\definecolor{ourred}{RGB}{231, 76, 60}
\definecolor{ourredborder}{RGB}{192, 57, 43}
\definecolor{ouryellow}{RGB}{241, 196, 15}
\definecolor{ouryellowborder}{RGB}{243, 156, 18}
\definecolor{ourpurple}{RGB}{155, 89, 182}
\definecolor{ourpurpleborder}{RGB}{142, 68, 173}
\definecolor{ourturquoise}{RGB}{26, 188, 156}
\definecolor{ourturquoiseborder}{RGB}{22, 160, 133}
\definecolor{ourturquoise}{RGB}{26, 188, 156}
\definecolor{ourturquoiseborder}{RGB}{22, 160, 133}
\definecolor{ourwhite}{RGB}{236, 240, 241}
\definecolor{ourwhiteborder}{RGB}{189, 195, 199}
\definecolor{ourgray}{RGB}{149, 165, 166}
\definecolor{ourgrayborder}{RGB}{127, 140, 141}

\definecolor{ourwhite2}{RGB}{246, 247, 248}

\definecolor{matplotlibblue}{HTML}{1f77b4}
\definecolor{matplotliborange}{HTML}{ff7f0e}
\definecolor{matplotlibgreen}{HTML}{2ca02c}

\newcommand{\backgroundcolorbox}[2]{\setlength{\fboxsep}{0em}{\hspace{-.3em}\rlap{\colorbox{#1}{\hspace{.3em}\phantom{#2}\hspace{.3em}}}\hspace{.3em}#2}}
\newcommand{\vone}[1]{\backgroundcolorbox{matplotlibblue!50}{\strut #1}}
\newcommand{\vtwo}[1]{\backgroundcolorbox{matplotlibblue!25}{\strut #1}}
\newcommand{\vthree}[1]{\backgroundcolorbox{matplotlibblue!10}{\strut #1}}

\newcommand{\shorttabular}[1]{\begin{tabular}{@{}c@{}}#1\end{tabular}}

\newcolumntype{H}{>{\setbox0=\hbox\bgroup}c<{\egroup}@{}}

\definecolor{cvprblue}{rgb}{0.21,0.49,0.74}
\usepackage[pagebackref,breaklinks,colorlinks,allcolors=cvprblue]{hyperref}

\def\paperID{31337} %
\def\confName{CVPR}
\def\confYear{2026}

\title{Envisioning the Future, One Step at a Time}

\author{
    Stefan Andreas Baumann\thanks{Equal Contribution.}\ \ $^{1,2}$ \quad Jannik Wiese\footnotemark[1]\ \ $^{1,2}$\\ Tommaso Martorella$^{1,2}$ \quad
    Mahdi M.~Kalayeh$^3$ \quad Bj\"orn Ommer$^{1,2}$\\[0.5em]
    $^1$CompVis @ LMU Munich \quad $^2$Munich Center for Machine Learning (MCML)
    \quad
    $^3$Netflix
}

\begin{document}

\maketitle

\begin{strip}
\vspace{-3.5em}
\centering
{
\includegraphics[width=\textwidth]{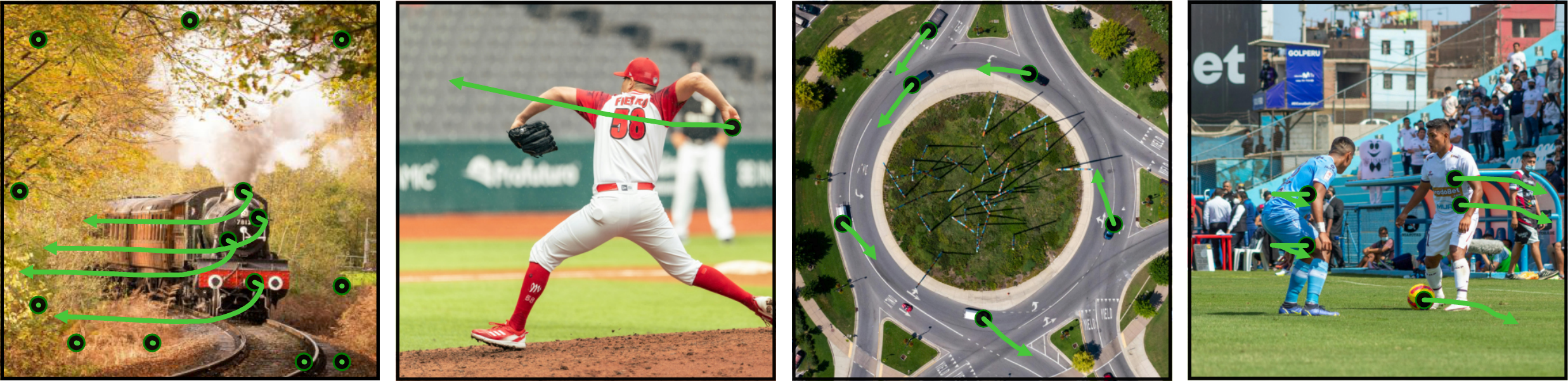}\vspace{.4em}
\includegraphics[width=\textwidth]{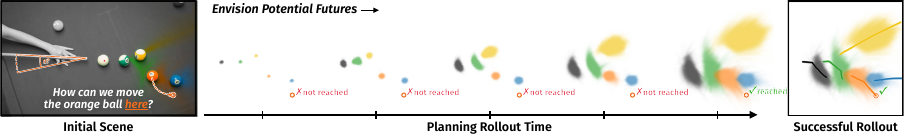}
}
\vspace{-1em}
\captionof{figure}{From a single image, our model envisions diverse, physically consistent futures in open-set environments (top). By exploring directly in motion space, it can rapidly perform thousands of counterfactual rollouts -- here to select a candidate billiard shot (bottom).}
\label{fig:teaser}
\end{strip}

\begin{abstract}
Accurately anticipating how complex, diverse scenes will evolve requires models that represent uncertainty, simulate along extended interaction chains, and efficiently explore many plausible futures. Yet most existing approaches rely on dense video or latent-space prediction, expending substantial capacity on dense appearance rather than on the underlying sparse trajectories of points in the scene. This makes large-scale exploration of future hypotheses costly and limits performance when long-horizon, multi-modal motion is essential.
We address this by formulating the prediction of open-set future scene dynamics as step-wise inference over sparse point trajectories. Our autoregressive diffusion model advances these trajectories through short, locally predictable transitions, explicitly modeling the growth of uncertainty over time. This dynamics-centric representation enables fast rollout of thousands of diverse futures from a single image, optionally guided by initial constraints on motion, while maintaining physical plausibility and long-range coherence.
We further introduce OWM, a benchmark for open-set motion prediction based on diverse in-the-wild videos, to evaluate accuracy and variability of predicted trajectory distributions under real-world uncertainty. Our method matches or surpasses dense simulators in predictive accuracy while achieving orders-of-magnitude higher sampling speed, making open-set future prediction both scalable and practical.\\
Project page: \href{http://compvis.github.io/myriad}{http://compvis.github.io/myriad}.
\end{abstract}

\section{Introduction}
\label{sec:intro}
A key feature of intelligence is the ability to \emph{envision} possible futures and use them to guide behavior~\cite{schacter2008episodic,schacter2017episodic,seligman2013navigating,schacter2007remembering}, rather than merely reacting after they have become reality -- anticipating how motion might unfold~\cite{battaglia2013simulation,keller2018predictive,ullman2017mind} rather than retracing how it already has. Since we live in a highly dynamic world, we need to quickly predict and simulate potential \emph{future} movements and interactions in the environment around us.
Yet, the complexity of our world is staggering:
every hidden contact, every subtle interaction could, in principle, dramatically change future scene dynamics.
Our minds cope with this open-set chaos through abstraction~\cite{johansson1973visual,blake2007perception}: we do not ``paint'' a picture of the future, we trace only the important changes that matter.
This sparsity is what makes efficiently envisioning the future possible, as long as it remains the future.

In contrast, most current generative (world) models, however, attempt the opposite. Video~\cite{brooks2024video,parkerholder2024genie2,ball2025genie3} and latent space~\cite{hafner2025training,zhou2025dinowm,karypidis2024dinoforesight} simulators predict dense representations of entire scenes, expending enormous capacity on aspects irrelevant to scene dynamics.
This makes envisioning the future in open-ended settings, precisely when many possible futures must be considered, prohibitively costly.

Moreover, the world is deeply interwoven and stochastic: between now and any future moment lies an immense chain of interactions and entanglements. Thus, predicting what the world will look like even a few seconds from now cannot be done in a single leap. Instead, we must simulate the intervening interactions step by step -- just as we do not foresee the outcome of billiard shot all at once, but unroll it gradually and abstractly, collision by collision.
Previous models that tried to predict that distant outcome in one step~\cite{baumann2025whatif} must implicitly account for every interaction at once. This implies an impossible burden unless tasks are trivially ``one-hop'' or model capacity is unbounded. Otherwise, the only feasible approach is to unfold the future step by step, progressing through short, locally predictable transitions where the web of interactions remains manageable.

Each step depends on the previous and models the growth of uncertainty over time.
This incremental structure is what makes reasoning under real-world complexity feasible.
Implementing this principle computationally allows us to envision the future, and the many ways of getting there, not just once, but thousands of times, effectively simulating the inherent stochasticity of our environment.

Technically, we formulate this as an autoregressive diffusion model over sparse trajectories.
It learns from diverse in-the-wild videos and generalizes to open-set dynamics of everyday scenes.
The model perceives the world through a single image and subsequently envisions diverse futures through fast rollouts, optionally guided by initial motion cues.
The efficient sparse representation of scene dynamics, rather than appearance, allows us to enumerate hypotheses and capture the stochasticity of our world orders of magnitude faster than dense video models.

To ground this task, we introduce \ourdataname, a benchmark for open-world motion prediction that evaluates whether models can generate physically consistent, diverse trajectories under real-world uncertainty.
Across both structured and open-set domains, our model achieves accuracy on par with or surpassing dense approaches, while enabling exploration of far more futures within the same compute budget.

By focusing on dynamics instead of pixels, we make motion prediction not only faster, but fundamentally more scalable: a model that does not paint the world frame by frame, but \emph{envisions how it moves}.

We summarize our main contributions as follows:
\begin{itemize}
    \item We cast visual motion prediction as \emph{open-set}, \emph{step-wise} modeling of distributions over \emph{sparse point trajectories} from a single image, allowing models to envision how complex, unconstrained scenes evolve without rendering appearance.
    \item We introduce an autoregressive diffusion model tailored to this formulation, with an efficiency-optimized architecture that enables large-scale, fast sampling of diverse futures.
    \item We present \ourdataname, a benchmark designed to evaluate the physical plausibility and accuracy of trajectory distributions under open-set conditions.
    \item We demonstrate that our approach matches or surpasses dense models in accuracy while being orders of magnitude faster, thereby enabling the exploration of thousands of plausible futures within the same compute budget.
\end{itemize}

\section{Related Work}
\label{sec:related_work}
We can examine the relevant literature on motion prediction from four distinct perspectives: \emph{visual tax}, \emph{granularity}, \emph{domain}, and \emph{paradigm}. A model that requires video generation as a prerequisite for motion prediction is considered \emph{dense} and incurs the \emph{visual tax}, as it must generate every pixel before it can reason about motion dynamics. \emph{Domain} refers to the environment in which a model operates and its ability to generalize to previously unobserved settings. For instance, a physics simulator may not incur the \emph{visual tax} because, after interpreting the scene, it relies solely on physics engines to reason about possible futures. However, such models often suffer from a limited \emph{domain}, rendering them obsolete or irrelevant in real-world and in-the-wild scenarios. Finally, \emph{paradigm} pertains to whether motion is modeled in a \emph{single-shot} or \emph{step-by-step} manner. The latter enables more sophisticated reasoning, allowing not only for the prediction of the final state but also for the explanation of motion dynamics (i.e., \emph{how} the system evolves to that state). In the remainder of this section, we adopt these definitions to briefly review the literature and clearly position our work relative to prior approaches.

Generation of potential motion from static images has been widely explored in the literature. Modern video generation~\cite{blattmann2023stable,yang2024cogvideox,chen2025skyreelsv2infinitelengthfilmgenerative,teng2025magi,brooks2024video,deepming2025veo3,openai2025sora2,runway2025gen4,pika2025pika21,kuaishou2024kling} and video world models~\cite{ball2025genie3,valevski2024diffusionmodelsrealtimegame,alonso2024diffusion,cheng2025playing,decart2024oasis,guo2025mineworld,seid2024lucidv1,jiang2025enerverse,zhu2025unified,bartoccioni2025vavim,hafner2025training,blattmann2021ipoke,blattmann2021understanding,dorkenwald2021stochastic,li2024puppet,venkatesh2024understanding} can produce \emph{dense} sequences of possible futures from a single starting image and/or a short context. However, these approaches incur a significant \emph{visual tax}: they model appearance and its temporal evolution alongside the \emph{dense} motion dynamics of the entire scene, making open-ended prediction, and especially branching, extremely expensive. Image-to-dense motion techniques~\cite{shi2024motion,liang2024movideo,zhou2025probdiffflow,bharadhwaj2024track2act,walker2015dense,li2024generative,boduljak2025happens} primarily aim to produce motion. When directly generating motion~\cite{shi2024motion,liang2024movideo,bharadhwaj2024track2act,li2024generative}, these methods can avoid the \emph{visual tax}. Nevertheless, by modeling \emph{all} motion rather than a decision-centric subset, they significantly increase computational demands for prediction and are prone to error accumulation. The same limitation applies to feature-space world models that operate on generic representations~\cite{karypidis2024dinoforesight,zhou2025dinowm,baldassarre2025back} or domain-specific image embeddings~\cite{ha2018world,hafner2019dream,hafner2020mastering,hafner2019learning,hafner2023mastering,bruce2024genie}. In contrast, our approach not only completely avoids the \emph{visual tax}, but also focuses computation \emph{only} on understanding motion by modeling distributions over a \emph{sparse} set of user-defined points. This eliminates the need for dense prediction of motion dynamics and enables extensive exploration of potential motion, including branching.

Another group of prior works first estimate the physical properties (e.g., object shape, mass, friction, pose) of the scene and then leverage off-the-shelf physics engines to predict scene motion~\cite{wu2015galileo,wu2017learning,jaques2019physics,battaglia2013simulation,liu2024physgen,chen2025physgen3dcraftingminiatureinteractive,xie2024physgaussian,li2025wonderplay,mottaghi2016newtonian}. These methods can produce highly accurate motion when the dynamics are fully in-domain for the physics engine and parameter estimation is exact, but they fail to generalize to truly open-set motion, including everyday scenarios or in-the-wild visuals. In contrast, our approach performs motion prediction in a fully open-set regime and learns all dynamics in a purely data-driven manner, without relying on external components such as a physics engine.

Most existing literature~\cite{gao2018im2flow,rosello2016predicting,pintea2018dejavumotionprediction,shin2024instantdrag,baumann2025whatif,vondrick2016anticipating} frames the problem of motion prediction from a single image as a one-shot task. These approaches either demand extremely high model capacity to handle multi-contact and long-horizon scenarios, or they incur a substantial \emph{visual tax}, comparable to that seen in auto-regressive video models. This limitation arises because such methods depend on pixel-level outputs to reason across multiple steps. In essence, after making a single-step prediction, the model must convert this prediction back into the visual domain~\cite{rosello2016predicting} before it can be used as input for generating the next step, resulting in a back-and-forth (i.e. encoding-decoding) process between real and latent spaces. In contrast, we employ \emph{step-wise} auto-regressive generation over \emph{sparse} points, enabling long horizons and explicit explorations. In other words, we demonstrate that multi-step reasoning about a scene's motion does not require attention to every single pixel, a property that unlocks significant potential for efficient, long-horizon, multi-step reasoning.  

It is worth mentioning that prior efforts exist in predicting the motion of a sparse set of objects; however, these methods typically operate in narrow domains such as multi-agent/social forecasting~\cite{alahi2016social,gupta2018social,salzmann2020trajectron++,ngiam2021scene}, autonomous driving~\cite{gao2020vectornet,liang2020learning,chai2019multipath,varadarajan2022multipath++,zhao2021tnt,nayakanti2022wayformer}, human-pose motion~\cite{yuan2023physdiff,bringer2024mdmp,ruizponce2025mixermdmlearnablecompositionhuman}, or fully specified custom environments~\cite{fragkiadaki2016visual,mottaghi2016happens}, and typically require abstract inputs, thereby limiting their general applicability. Unlike these works, we specifically target \emph{open-set} motion prediction in unconstrained scenes at a granularity specified by the user during inference, learning to parse and reason directly in a multi-step manner from appearance at sparse decision points.

In summary, compared to prior work, our approach offers key advantages across all four axes defined at the outset. Unlike dense video generation and feature-based models, which pay a high \emph{visual tax} by operating at the pixel level, our method entirely avoids this cost by modeling motion only over a \emph{sparse} set of user-defined points, thus achieving fine control over \emph{granularity}. In terms of \emph{domain}, whereas physics-based and domain-specific models are limited to narrow or closed environments, our data-driven approach generalizes to open-set, unconstrained scenes. Finally, rather than relying on a \emph{single-shot} paradigm, we employ \emph{step-wise} auto-regressive reasoning, enabling efficient, interpretable, and long-horizon motion prediction, including branching, without the need for dense reconstruction at each step. This combination of low visual tax, user-controlled granularity, open-set domain coverage, and step-wise paradigm distinguishes our method from the existing literature.

\section{Methodology}
\label{sec:methodology}

We consider a single reference frame $\mathcal{I}_0$ at time $t = 0$.
Given a sparse set of $K$ visible query points ${\mathbf{x}_0 := \{x_0\supi\}_{i=1}^K}$, with $x_t\supi \in \mathbb{R}^2$, the goal is to model a distribution over their {full} future trajectories
\begin{equation}\label{eq:problem_statement_distribution_target}
    p(\underbrace{\mathbf{x}_{t=1}, \mathbf{x}_{t=2}, \ldots, \mathbf{x}_{t=T}}_{\mathrel{=:}\ \mathbf{x}_{1:T}} \mid \mathbf{x}_{0}, \mathcal{I}_{0}),
\end{equation}
in the same 2D reference frame, assuming a static camera.
This joint distribution captures the independent evolution of trajectories, their interactions, and their interdependencies.
We model incremental motion at each timestep ${\Delta x_{t}\supi := x_{t + 1}\supi - x_{t}\supi}$, with trajectories obtained by accumulating increments over time starting from $\mathbf{x}_0$.
{Optionally, an initial motion hint (``poke'') $\Delta x_0\supi$ can be provided as conditioning to guide the predicted trajectories.}

\paragraph{Autoregressive Formulation}
We parametrize the joint with an autoregressive transformer~\cite{vaswani2017attention}~$p_\theta$, factorizing causally over time and, within each step, over trajectories, as%
\begin{equation}
\begin{aligned}
    &p_\theta(\mathbf{x}_{1:T} \mid \mathbf{x}_{0}, \mathcal{I}_{0})\\
    &= \prod_{t=1}^T p_\theta(\mathbf{x}_{t} \mid \mathbf{x}_{<t}, \mathcal{I}_{0}) &&\!{\color{ourgrayborder}\triangleright\ \text{Time}}\\
    &= \prod_{t=1}^T \prod_{i=1}^K p_\theta(x_{t}\supi \mid \mathbf{x}_{t}\supi[<i], \mathbf{x}_{<t}, \mathcal{I}_{0}). &&\!{\color{ourgrayborder}\triangleright\ \text{Trajectories}}\!\! \label{eq:full_factorization}
\end{aligned}
\end{equation}
This factorization reflects how humans often reason step by step temporally~\cite{zacks2007event,battaglia2013simulation,forstmann2016sequential} and makes the interdependence between trajectories explicit by conditioning each update on all previously realized points at the current time and the full past.
Importantly, this formulation enables fast decoding with KV caching.
In practice, the model predicts $\Delta x_{t}\supi$ and updates $x_t\supi$ online.
We encode the image $\mathcal{I}_0$ into spatial features $\mathbf{E}_\mathrm{img}$ via an encoder~\cite{dosovitskiy2021an} ${\mathcal{E}_\psi}$ with parameters $\psi$.

\begin{figure}[t]
    \centering
    \includegraphics[scale=1]{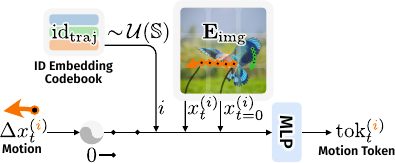}
    \caption{\textbf{Motion Token Construction.} The fourier-embedded motion $\Delta x_t\supi$ (alternatively a zero-vector) is combined with a per-trajectory unique randomized trajectory identifier $\mathrm{id}_\mathrm{traj}\supi$ and the local image features, retrieved at the current $x_t\supi$ and original position $x_{t=0}\supi$, providing information about what it is and local context.}
    \label{fig:motion_token_enc}
\end{figure}

\paragraph{Motion Tokens}
Each motion token corresponds to a particular $(t, i)$ pair and aggregates three kinds of information.
First, we retrieve appearance (``what'') from the spatial image features $\mathbf{E}_\mathrm{img}$ at the trajectory's \emph{origin} $x_0\supi$ using bilinear sampling.
Similarly, we retrieve local context (``where'') from the features at the \emph{current} position $x_t\supi$.
Second, we encode current motion $\Delta x_t\supi$ as a Fourier embedding~\cite{mildenhall2020nerf,tancik2020fourfeat} when observed; for query tokens, we substitute a zero vector of the same dimension.
Third, we encode identity (``who'') by a trajectory-specific vector $\mathrm{id}_\mathrm{traj}\supi \in \mathbb{R}^d${, which we find to be critical for successful modeling in multi-trajectory settings}.
Rather than using a finite codebook, we draw $\mathrm{id}_\mathrm{traj}\supi \sim \mathcal{U}(\mathbb{S}^{d-1})$ (the unit sphere in $\mathbb{R}^d$) each iteration.
Random unit-sphere directions yield nearly orthogonal IDs, scale to arbitrary $K$, and prevent the model from becoming overly reliant on specific indices.
We fuse these three sources into the motion token $\mathrm{tok}_t\supi \in \mathbb{R}^{d_\mathrm{model}}$ using a small MLP.
We show an illustration of the whole mechanism in \cref{fig:motion_token_enc}.

\definecolor{pe_current}{RGB}{249,196,143}
\definecolor{pe_origin}{RGB}{155,211,151}
\definecolor{pe_t}{RGB}{144,179,225}
\begin{figure}[t]
    \centering
    \includegraphics[scale=1]{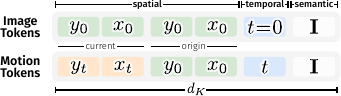}
    \caption{\textbf{Positional Encoding Scheme.} We encode the {\color{pe_current}{current}} and {\color{pe_origin}{original}} spatial position of each token, alongside its {\color{pe_t}{time}}. Motion tokens attend to each other and to image tokens.}
    \vspace{-.5em}
    \label{fig:positional_encoding}
\end{figure}

\paragraph{Shared Spatiotemporal Positional Encoding}
Motion and image tokens share one reference coordinate frame, so we apply a single positional encoding scheme to both.
We base our positional encoding on axial RoPE~\cite{su2021roformer,crowson2024hourglass}.
Each motion token receives spatial encodings for the \emph{current} position $x_t\supi$, the \emph{origin} $x_0\supi$, plus time $t$.
Image tokens use the same position at $t = 0$ for both 2D position slots.
This way, motion tokens can attend to both context about them (``what'') at their original location, and local context (``where'') at their current position.
Finally, we reserve a slice of channels without positional encoding to enable global (semantic) attention~\cite{barbero2024round,crowson2024hourglass}.
We illustrate the layout in \cref{fig:positional_encoding}.

\begin{figure}[t]
    \centering
    \adjustbox{max width=\linewidth}{
        \includegraphics[scale=1]{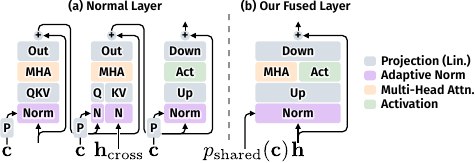}
    }
    \caption{\textbf{Fast Reasoning Blocks.} \textbf{(a)} Previous methods~\cite[cf.][]{baumann2025whatif} use normal transformer layers, incurring significant overhead due to the multitude of operations performed per block. \textbf{(b)} Our fused layers reduce complexity significantly, improving efficiency.}
    \vspace{-.5em}
    \label{fig:fused_layer_comparison}
\end{figure}

\paragraph{Fast Reasoning Blocks}
We aim to explore a multitude of motion hypotheses efficiently, so we design the backbone for high rollout throughput.
Instead of evolving the hidden state $\mathbf{h}$ using normal sequential transformer layers, i.e.,%
\begin{equation*}
\begin{aligned}
    &\mathbf{h} \gets \mathbf{h} + \mathrm{SA}(\mathbf{h}), &&{\color{ourgrayborder}\triangleright\ \text{Self-Attention}}\\
    &\mathbf{h} \gets \mathbf{h} + \mathrm{CA}(\mathbf{h}, \mathbf{h}_\mathrm{cross}), \quad &&{\color{ourgrayborder}\triangleright\ \text{Cross-Attention}}\\
    &\mathbf{h} \gets \mathbf{h} + \mathrm{FFN}(\mathbf{h}), &&{\color{ourgrayborder}\triangleright\ \text{Feedforward Network}}
\end{aligned}
\end{equation*}
we adopt parallel transformer blocks~\cite{gptj} with one residual:%
\begin{equation}
    \mathbf{h} \gets \mathbf{h} + \mathrm{SA}(\mathbf{h}) + \mathrm{CA}(\mathbf{h}, \mathbf{h}_\mathrm{cross}) + \mathrm{FFN}(\mathbf{h}).
\end{equation}
\begin{wrapfigure}{r}{.2\linewidth}
    \vspace{-1.1em}
    \captionsetup[figure]{margin={-1.5em,0em}}
    \hspace{-.36\linewidth}\includegraphics[width=1.35\linewidth]{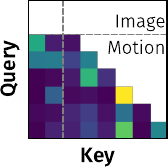}
    \captionof{figure}{Our attention mask.}
    \vspace{-2em}
\end{wrapfigure}
We share pre-normalization and fuse projections such that one ``up'' computes QKV and FFN-up, and one ``down'' merges attention and FFN outputs.
Further, we combine self- and cross-attention in a prefix layout, concatenating $[\mathbf{h}_\mathrm{image}|\mathbf{h}_\mathrm{motion}]$ and masking such that image tokens attend to nothing (emulating cross-attention, unlike previous approaches~\cite{raffel2020exploring,esser2024scaling}, that modify these tokens over depth) and motion tokens attend (causally) to both streams.
This cuts down kernel launches significantly.
The final fused step becomes
\begin{equation}
    \!\!\!\mathbf{h} \gets \mathbf{h} + \mathrm{Down} \circ \mathrm{\Bigl[\frac{MHA}{Act}\Bigr]} \circ \mathrm{Up} \circ \mathrm{Norm}(\mathbf{h}, p_\mathrm{shared}(\mathbf{c})),
\end{equation}
with conditioning implemented via adaptive norms~\cite{huang2017arbitrary} with a shared~\cite{chen2024pixartalpha} control vector $p_\mathrm{shared}(\mathbf{c})$, mapping the (optional) model condition $\mathbf{c}$.
We show a comparison of our blocks with a typical layer structure in \cref{fig:fused_layer_comparison}.

\begin{figure}[t]
    \centering
    \adjustbox{max width=\linewidth}{
        \includegraphics[scale=1]{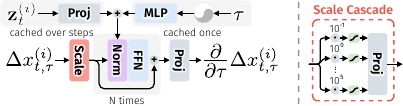}
    }
    \caption{\textbf{Posterior FM Head.} \textit{Left:} Our FM Head consists of multiple FFN blocks conditioned on $\mathbf{z}_t\supi$ and flow matching time $\tau$ via adaptive norms~\cite{huang2017arbitrary}. We set up the conditioning mechanism such that every component can be cached, reducing computations. \textit{Right:} Our multiscale, tanh-saturated input stack helps stabilize behavior when modeling motion with heavy-tailed behavior.}
    \label{fig:flow_matching_head}
\end{figure}

\paragraph{Posterior Parametrization with Flow Matching (FM)}
We parametrize the conditional in \cref{eq:full_factorization} as a distribution over stepwise motion $\Delta x_t\supi$. A flow matching~\cite{lipman2022flow} head~\cite[cf.][]{li2024autoregressive} $v_\phi$ predicts the ODE velocity of a noisy motion $\Delta x_{t, \tau}\supi$ as it evolves from $\tau = 0$ (Gaussian prior) to $\tau = 1$ (data):
\begin{equation}
    v_\phi: (\Delta x\supi_{t,\tau}, \tau, \mathbf{z}_t\supi) \mapsto \frac{\partial }{\partial \tau} \Delta x\supi_{t,\tau},
\end{equation}
with parameters $\phi$.
The AR backbone maps the conditioning to a compact representation $\mathbf{z}_t\supi$ that conditions the head.
We set up the head architecture such that separate branches encode $\tau$ and $\mathbf{z}_t\supi$ (see \cref{fig:flow_matching_head}), enabling caching instead of recomputation at every sampling step.
{Compared to parametrizing the distribution using GMMs~\cite{baumann2025whatif,tschannen2024givt}, we find that this leads to both significantly faster convergence during training and significantly more accurate predictions.}

\begin{wrapfigure}{r}{.2\linewidth}
    \vspace{-1em}
    \captionsetup[figure]{margin={-1.5em,0cm}}
    \hspace{-.36\linewidth}\includegraphics[width=1.35\linewidth]{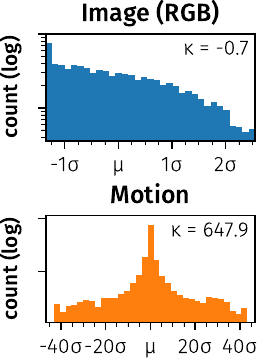}\vspace{-.3em}
    \captionof{figure}{Value distribution.}
    \vspace{-1.5em}
\end{wrapfigure}
\paragraph{Scale Cascade}
Motion shows significant heavy tail-like behavior, unlike typical image distributions for which similar heads were previously applied~\cite{li2024autoregressive,fan2025fluid}, with excess kurtosis $\kappa$ in the hundreds instead of around 0.
We account for this using a high-variance noise prior, setting $\sigma_\mathrm{noise} \gg \sigma_\mathrm{data}$, and help the head deal with the large range of value scales present on the input side.
Specifically, we create a cascade of logarithmically spaced scale coefficients $\mathbf{s}$ and feed $\tanh(\mathbf{s} \cdot \Delta x_{t,\tau}\supi)$ component-wise to the head, where small scales preserve fine motion detail while large scales saturate, bounding the influence of rare extremes (see \cref{fig:flow_matching_head}, right).
This gives the network stable features for tiny motions and large jumps at once, without letting outliers dominate.

\paragraph{Objective and Training}
We train with teacher forcing~\cite{baumann2025whatif} and maximize the likelihood in \cref{eq:full_factorization} through the augmented ELBO defined by the flow matching loss~\cite{lipman2022flow}
\begin{equation}
    \label{eq:our_fm_loss}
    \mathcal{L}_\text{FM} = \!\!\!\!\!\! \mathop{{}\mathbb{E}}_{\tau, \Delta x_{t,0}\supi, \Delta x_{t,1}\supi}\|v_\phi(\Delta x_{t,\tau}\supi | \mathbf{z}_t\supi) + \Delta x_{t,0}\supi - \Delta x_{t,1}\supi\|_2^2.
\end{equation}
We train the FM head $v_\phi$, AR transformer $p_\theta$, and image encoder $\mathcal{E}_\psi$ end-to-end, jointly optimizing $(\theta, \psi, \phi)$.
{Supervision is obtained from videos with (pseudo-)ground truth trajectories obtained, e.g., from off-the-shelf trackers~\cite{zholus2025tapnext,karaev2024cotracker3}.}

\paragraph{Inference}
We decode step by step with KV caching following the factorization in \cref{eq:full_factorization}.
For each $(i, t)$, the transformer predicts $p_\theta(\Delta x_{t}\supi \!\mid\! \mathbf{x}_{t}\supi[<i]\!\!, \mathbf{x}_{<t}, \mathcal{I}_{0})$ via $\mathbf{z}_t\supi$.
Sampling $\Delta x_{t}\supi$ is done by solving the ODE defined by $v_\phi(\ \cdot \mid \mathbf{z}_t\supi)$.

\section{Benchmarking Efficient Open-World Motion Prediction}\label{sec:benchmark}
Open-world scenes are {messy}, and ambiguous, but, more importantly, realized only once as we only ever observe a single future.
Therefore, to properly evaluate open-world motion prediction, one must assess not a point estimate, but rather the \emph{distribution} of all feasible trajectories that are consistent with the observed future.
To make such distribution evaluations feasible, given only a single ground truth observation, the distribution of plausible motion has to be limited in complexity.
To this end, we curate a diverse open-world benchmark dataset for motion prediction under a static-camera assumption to remove viewpoint confounders.

\subsection{Data}
\begin{figure}
    \centering
    \includegraphics[width=\linewidth]{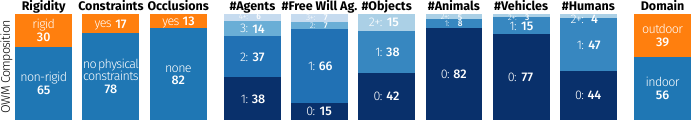}{\vspace{.3em}}
    \includegraphics[width=\linewidth]{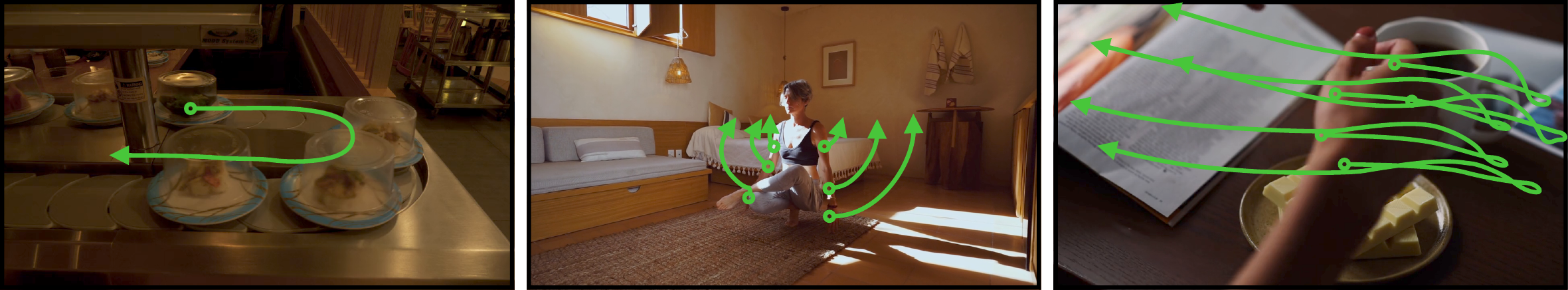}
    \caption{\textbf{OWM Composition.} We curate OWM to cover a wide variety of settings. \textit{Top:} dataset statistics. \textit{Bottom:} some examples.}
    \label{fig:owm_composition}
\end{figure}

\paragraph{\ourdataname}
We curate a set of 95 diverse in-the-wild videos selected for varied motion dynamics.
For each scene, we provide a reference frame $\mathcal{I}_{0}$, query points $\mathbf{x}_0$ with the observed ground truth motion $\mathbf{x}_{1:T}$ for a duration between 2.5s and 6.5s (obtained using off-the-shelf trackers and verified to be accurate).
The cameras are verified to be static to enable objective evaluation of predicted scene motion.
We show composition statistics in \cref{fig:owm_composition}.
\ourdataname is solely used for evaluation and {will be made publicly available}.

\paragraph{Physical Diagnostics Sets}
We supplement \ourdataname with two additional sets of videos in more constrained settings, focusing on simple physics motion principles.
We source these sets from PhysicsIQ~\cite{motamed2025generative}, specifically the ``solid mechanics'' subset, and Physion~\cite{bear2021physion}, and manually annotate reference frames and query points consistent with \ourdataname.

\subsection{Efficient Motion Hypothesis Generation}
\paragraph{Task}
{Given a single RGB input image $\mathcal{I}_{0}$ and a short warm-up hint $h_0$ (the motion over the first 2 frames $\mathbf{x}_{0:2}$), predict a \emph{set of future trajectory samples} for provided query points $\mathbf{x}_0$ over timesteps $t = 1, \ldots, T$.
When evaluating video generation models, we provide the hint as full additional frames and obtain trajectories from generated videos using off-the-shelf point trackers.}

\paragraph{Hypothesis Generation}
We report results under two standardized budgets:
\begin{enumerate}
    \item \textbf{Best-of-$N$.} Sample $N = 5$ sets of trajectories, and evaluate the closest to the ground truth observation.
    \item \textbf{Best-within-Timelimit (primary).} Allocating fixed wall-clock on a reference GPU per scene (5min on an Nvidia {H200} to enable the evaluation of video models), methods may generate \emph{any number of hypotheses}, which are subsequently evaluated following the \emph{Best-of-$N$} setting. This setting enables measuring \emph{search efficiency}.
\end{enumerate}
Further implementation details are specified in the appendix.

\paragraph{Metrics}
From the multiple generated hypotheses, we compute prediction error via the pointwise distance of each predicted trajectory $\{\hat{\mathbf{x}}_{n,1:T}\}_{n=1}^N$ with the ground truth observation $\mathbf{x}_{1:T}$, using the mean distance over the prediction horizon $T$ for the closest trajectory
\begin{equation}
    \mathrm{minADE}_N = \min_k \Bigl[\frac{1}{KT}\sum_{i=1}^K\sum_{t=1}^T\bigl\|\hat{\mathbf{x}}_{n,t}\supi - \mathbf{x}_t\supi\bigr\|_2^2\Bigr],
\end{equation}
akin to a one-sided Wasserstein distance over motion space. This captures whether the distribution covers the true outcome without penalizing alternative plausible futures.

\section{Experiments}
\label{sec:experiments}

\subsection{Implementation Details}
We use L-scale transformers~\cite{vaswani2017attention} for both the {motion model} and the image encoder, the latter of which we initialize with DINOv3-L/16~\cite{simeoni2025dinov3}, with input resolution $512^2$.
Our flow matching head shares its width with the {motion model} and has a depth of 3. In total, we have 665M trainable parameters.
We train using bfloat16 mixed precision with AdamW~\cite{loshchilov2018decoupled,kingma_adam_2017} with a peak learning rate of 3e-5, betas (0.9, 0.99), and weight decay $0.01$.
The learning rate is linearly warmed up over the first 5k steps with subsequent linear decay to 1e-8.
In general, we train with a global batch size of 128 scenes, using $K = 16$ trajectories and $T = 16$ timesteps for 400k steps, taking about 20 hours to converge on 16 Nvidia H200 GPUs.
We primarily train our models on a diverse dataset of 10M open-set video clips {collected from the internet}%
, with pseudo ground-truth motion obtained using TAPNext~\cite{zholus2025tapnext}.
{We additionally train a model using 3D tracks obtained using V-DPM~\cite{sucar2026vdpm}.
These tracks are then projected to the first camera view to induce a static camera, 
enabling direct learning of scene motion disentangled from camera motion.
These models are trained on a smaller subset of $\sim$1.5M clips due to the high cost of running such tracker models.}
For planning tests, we train separate models on data obtained from a billiard simulation~\cite{ebke2025pythonbilliards}.
Further details and ablations are in supplementary \cref{sec:appendix_additional_details,sec:appendix_additional_ablations}.

\subsection{Motion Prediction}

We evaluate our model's ability to predict motion in intricate real-world scenes using the {\ourdataname} dataset in \cref{tab:phys_correctness}a.
Using the same number of trials for all models in the \textbf{Best-of-5} setting, our approach generates a prediction that matches the observed motion with a higher degree of accuracy than state-of-the-art video generation models. Therefore, our approach is able to capture realistic motion more accurately than prior methods while being substantially faster and using significantly less parameters.
Under a constrained inference time budget in the \textbf{Best-within-5\,min}, our approach has a strong advantage due to its orders of magnitude better efficiency achieved by avoiding the visual tax of RGB world simulation, leading to a substantial widening of the accuracy gap.
In addition to \ourdataname, we further evaluate the physical understanding of our model on the PhysicsIQ~\cite{motamed2025generative} and Physion~\cite{bear2021physion} subsets in \cref{tab:phys_correctness}b-c.
Similar to the open-world setting, we find that our model is competitive with or outperforms state-of-the-art video models in the \textbf{Best-of-5} setting already, with the gap widening if time constraints are used.

\begin{figure}[t]
    \centering
    \includegraphics[width=\linewidth]{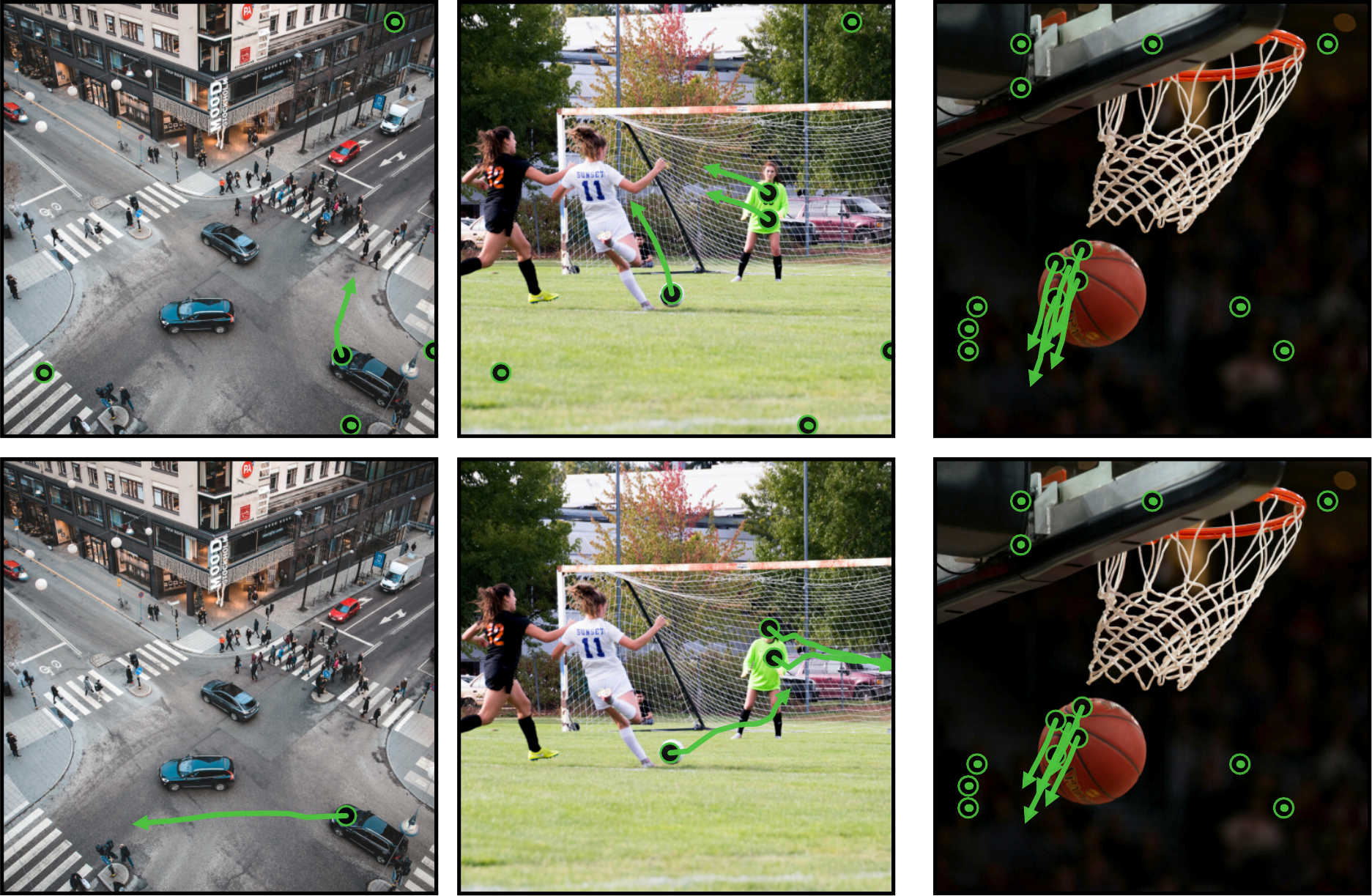}
    \begin{subfigure}{.65\linewidth}
        \centering
        \caption{Diverse Actions from a Single Image.}
        \label{fig:action_reasoning}
    \end{subfigure}\hfill
    \begin{subfigure}{.32\linewidth}
        \centering
        \caption{Object Coherence.}
        \label{fig:appearance_motion_placeholder}
    \end{subfigure}
    \caption{\textbf{(a)} Given different input pokes{\ (initial motion)}, our model produces different motions (visualized as {\color{ourgreenborder}\textbf{green}} lines) that adhere to constraints of the environment. \textbf{(b)} Our model predicts coherent motion for multiple points on the same object.}
    \vspace{-1em}
\end{figure}

Qualitative samples in \cref{fig:action_reasoning} show our model's capability to produce motion that is informed by visual cues in the scene. The motion rollouts respect constraints and adhere to specific kinematics of the objects visible in the scene. This also applies when predicting the motion of multiple points that move together in the context of the scene (see \cref{fig:appearance_motion_placeholder}).

\begin{table*}[t]
    \centering
    \begin{minipage}{.7201\linewidth}
        \adjustbox{max width=\linewidth}{
            \begin{tabular}{lccc@{\hskip .33em}cc@{\hskip .33em}cc@{\hskip .33em}c}
                \toprule
                \multirow{2}{*}[-3pt]{\textbf{Method}} & \multirow{2}{*}[-3pt]{\textbf{Param}} & \multirow{2}{*}[-3pt]{\begin{tabular}[c]{@{}c@{}}\textbf{Throughput} \\ (samples/min)\end{tabular} $\uparrow$} & \multicolumn{2}{c}{\textbf{(a) \ourdataname}} & \multicolumn{2}{c}{\textbf{(b) PhysicsIQ~\cite{motamed2025generative}}} & \multicolumn{2}{c}{\textbf{(c) Physion~\cite{bear2021physion}}} \\
                \cmidrule(lr){4-5} \cmidrule(lr){6-7} \cmidrule(lr){8-9}
                & & & \textsc{\footnotesize {Best-5$\downarrow$}} & \textsc{\footnotesize {Best-5min$\downarrow$}} & \textsc{\footnotesize {Best-5$\downarrow$}} & \textsc{\footnotesize {Best-5min$\downarrow$}} & \textsc{\footnotesize {Best-5$\downarrow$}} & \textsc{\footnotesize {Best-5min$\downarrow$}} \\
                \midrule
                MAGI-1~\cite{teng2025magi} & 4.5B       & 0.303 & \underline{0.037} & \underline{0.066} & 0.126 & 0.169 & \underline{0.061} & \underline{0.081}  \\
                Wan2.2~\cite{wan2025} & 14B        & 0.141 & 0.039 & DNF & 0.116 & DNF & 0.069 & DNF  \\
                CogVideo-X 1.5~\cite{yang2024cogvideox} & 5B & 0.051 & 0.051 & DNF & \textbf{0.100} & DNF & 0.063 & DNF  \\
                SkyReels V2~\cite{chen2025skyreelsv2infinitelengthfilmgenerative} & 1.3B  & 0.304 & 0.058 & 0.068 & 0.128 & \underline{0.137} & 0.069 & 0.084  \\
                SVD 1.1~\cite{blattmann2023stable} & 1.5B      & \underline{0.714} & 0.054 & 0.119 & 0.138 & 0.241 & 0.070 & 0.147  \\
                \midrule
                \modelname (Ours) & {665M} & {\textbf{2200}} & \textbf{0.029} & \textbf{0.013} & \underline{0.115} & \textbf{0.045} & \textbf{0.048} & \textbf{0.020} \\
                \color{ourgray} \modelname{\tiny Trained on 3$\rightarrow$2D Tracks} & \color{ourgray} {665M} & \color{ourgray} {2200} & \color{ourgray} 0.036 & \color{ourgray} 0.020 & \color{ourgray} 0.117 & \color{ourgray} 0.043 & \color{ourgray} 0.048 & \color{ourgray} 0.028 \\
                \bottomrule
            \end{tabular}
        }
        \caption{\textbf{Open-world \& Physical Motion Prediction.} We evaluate motion prediction capabilities across both open-world and constrained physical settings using the benchmark introduced in \cref{sec:benchmark}. Eliminating the need to model fine-grained pixel-level details lets our model focus on the dynamics of the scene, making it competitive with state-of-the-art video models in the Best-5 setting across all three subsets, despite having substantially fewer parameters and being substantially more efficient. The gap widens significantly in the efficiency-focused Best-5min setting, driven by the higher throughput.}
        \label{tab:phys_correctness}
    \end{minipage}\hfill%
    \begin{minipage}{.248\linewidth}
        \includegraphics[width=\linewidth]{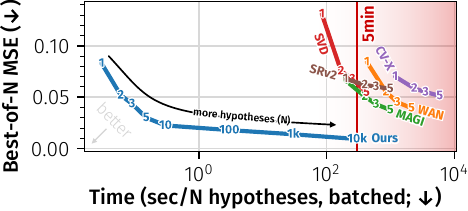}\vspace{.61em}
        \captionof{figure}{\textbf{Time-Accuracy Trade-off on OWM.} Higher numbers of hypotheses $N$ (denoted as numbers in lines) allow more accurate recovery of the observed motion. Across models, the relative improvement in accuracy with $N$ is comparable; the sparsity of our method makes it orders of magnitude more efficient.}
        \label{tab:owm_plots}
    \end{minipage}
    \vspace{-1em}
\end{table*}

\subsection{Action Selection by Envisioning Futures}

    We push beyond passive motion prediction and test whether the our model can be applied to choosing an action that leads to a desired outcome, in a fully zero-shot manner. {In billiard terms: can it plan a shot?}
    Unlike pure forward prediction compared to one observed future, this setting forces exploration of counterfactual futures -- many possible actions, many possible rollouts, one desired goal.

\begin{figure}[tb]
    \centering

    \includegraphics[width=\linewidth]{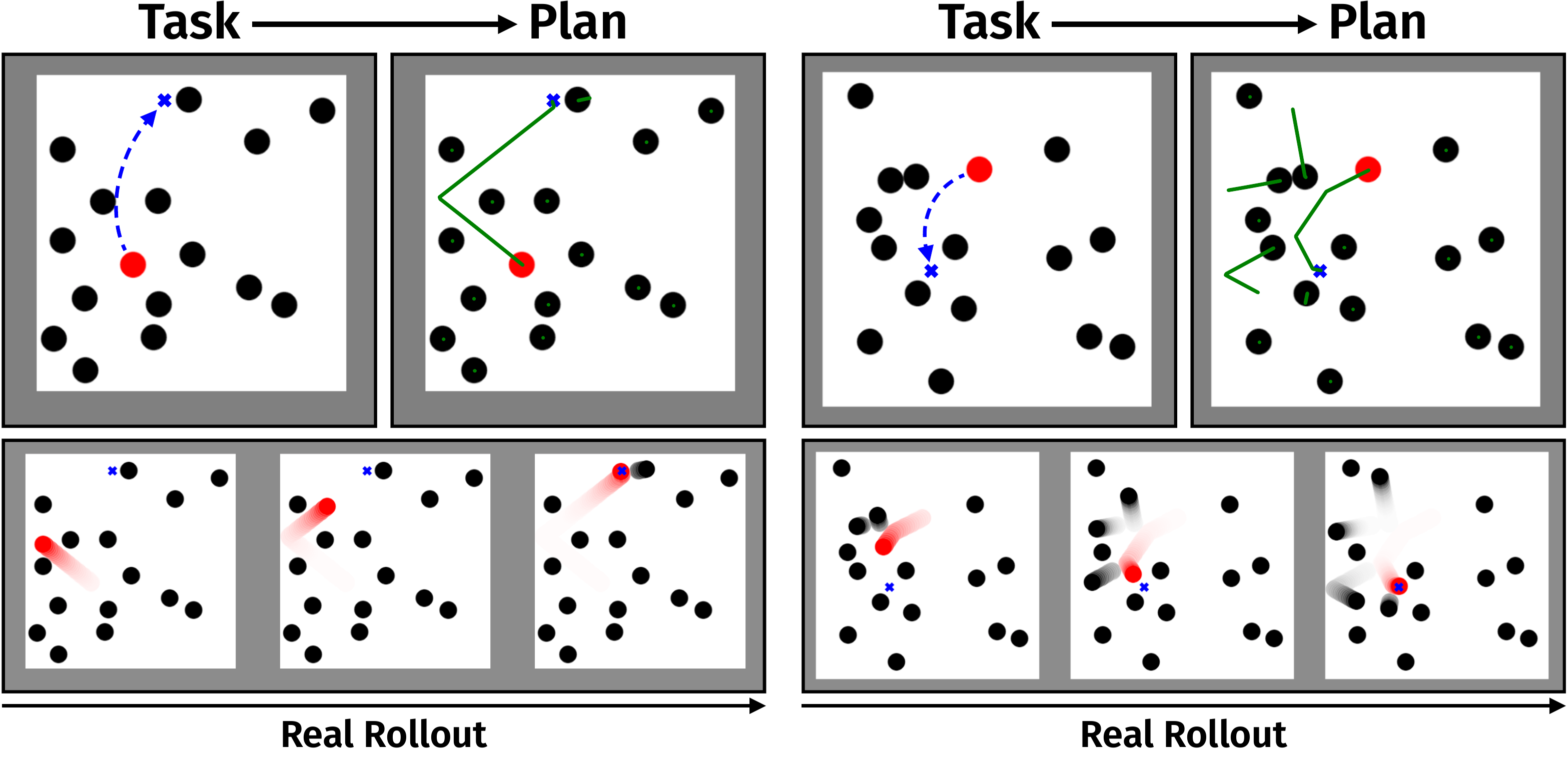}
    \caption{\textbf{Planning a Billiard Shot.} We search for a plan to move the red ball to the goal (top left). Our model derives a plan (top right) by predicting motion for different initial actions. Executing the action moves the ball to the desired location (bottom).}
    \label{fig:billiard_qual}
\end{figure}

    \paragraph{Setup}
    We use a billiard simulator~\cite{ebke2025pythonbilliards} to generate training data and evaluate all methods on an equal footing; every model is trained from scratch at a comparable scale. Each episode starts with a single image of the table, from which the model predicts future trajectories given the initial ball configuration $\mathbf{x}_0$ and an initial cue-ball impulse $\Delta x_0\supi[0]$.
    A ``plan'' constitutes selecting an initial strike direction and magnitude $a = (\theta, m)$.
    We sample a set of candidate actions $\{a_j\}$, predict the corresponding rollouts $\mathbf{x}_{j,1:T}$, and evaluate each rollout using a goal reward $R(\mathbf{x}_{j,1:T})$. 
    This is repeated until a time budget expires, then the plan that maximizes the expected reward is chosen and executed:
    \begin{equation}\label{eq:expected_reward}
        a^* = \arg \max_{a_j} \mathbb{E}_{\mathbf{x}_{1:T} \sim p_\theta(\ \cdot\ \mid\ \mathcal{I}_0, \mathbf{x}_{0}, a_j)}[R(\mathbf{x}_{1:T})].
    \end{equation}
    Performance is measured by the minimal $\ell_2$ distance between the target ball's location and the goal. {We calculate the accuracy of solving the task by thresholding the distance using the size of the ball.}
    We provide a visual explanation of the Billiard planning task in \cref{fig:billiard_qual}-top.

    \paragraph{Baselines}
    We compare against a wide range of baselines representative of common approaches. First, we compare with image-to-video generation methods, starting from an original image, with the initial cue ball impulse specified via either a second frame or a ``poke conditioning'' mechanism specifying the initial motion. We combine this with either full-sequence video diffusion, following standard video diffusion methods~\cite{blattmann2023stable,yang2024cogvideox,brooks2024video}, or framewise autoregressive video diffusion~\cite{teng2025magi,chen2025skyreelsv2infinitelengthfilmgenerative,chen2024diffusion,ruhe2024rollingdiffusionmodels}.
    We also include full-sequence trajectory diffusion~\cite[cf.][]{bharadhwaj2024track2act} and the flow poke transformer~\cite{baumann2025whatif}.

    \paragraph{Results}
    We show our findings in \cref{tab:billiard_acc}.
    Compared to image models, sparse trajectory models show at least an order of magnitude improvement in throughput, enabling higher accuracies.
    At the same time, directly ``leaping'' to the final state~\cite{baumann2025whatif}, while having the highest throughput, is not accurate enough to enable accurate predictions in such complex settings. Similarly, full trajectory diffusion~\cite{bharadhwaj2024track2act}, where the model does not gradually unroll the future step by step temporally but immediately has to denoise even steps far in the future, also significantly underperform. Our model combines both sparsity, enabling high throughput by forgoing the ``visual tax'', and step-by-step unrolling of the future, resulting in the highest accuracy.
    We also ablate regressing the next step instead of modeling the posterior. For a highly predictable environment like billiard, where little uncertainty is present, this also performs well, although it still underperforms to full distributional modeling. {Using a GMM posterior~\cite{baumann2025whatif,tschannen2024givt} is also worse than our FM head.}
    We visualize our method's planned actions in \cref{fig:billiard_qual}-bottom.

\begin{table}[tb]
    \centering
    \adjustbox{max width=.75\linewidth}{\centering
\begin{tabular}{l@{\hskip -.8em}c@{\hskip .35em}c} 
\toprule
\textbf{Method} & \textbf{Accuracy$\uparrow$} & \shorttabular{\textbf{Throughput}\\[-.3em]\footnotesize (actions/min)}$\uparrow$ \\

\midrule

{\color{ourgray} Simulator Oracle} %
& {\color{ourgray} 84\%}  %
& {\color{ourgray} 55,162.2}  %
\\

Image to Video Diff. (poke-cond.)~\cite[cf.][]{shi2024motion,liang2024movideo}   %
&  16\% %
&  \phantom{00,0}20.4 %
\\

Images to Video Diff.~\cite[cf.][]{chen2024diffusion,wan2025,yang2024cogvideox}
&  16\% %
&  \phantom{00,0}19.8 %
\\

AR Image to Video Diff. (poke-cond.)
&  12\% %
&  \phantom{00,0}22.2   %
\\

AR Images to Video Diff.~\cite[cf.][]{teng2025magi,chen2025skyreelsv2infinitelengthfilmgenerative,chen2024diffusion,ruhe2024rollingdiffusionmodels}   %
&  \phantom{0}8\% %
&  \phantom{00,0}18.6 %
\\

Full Trajectory Diffusion~\cite[cf.][]{bharadhwaj2024track2act}   %
&  \phantom{0}8\% %
&  \phantom{00,}160,8
\\

Flow Poke Transformer~\cite{baumann2025whatif}   %
&  \phantom{0}4\% %
&  \textbf{13,422.6} %
\\

\midrule

\modelname{\tiny Regression Head}   %
&  \underline{36\%}
&  \phantom{00,}\underline{754.6} %
\\

\modelname{\tiny GMM Head}   %
&  {24\%}
&  \phantom{00,}753.4
\\

\modelname (Ours)
&  \textbf{78}\rlap{\textbf{\%}}\phantom{\%}
& \phantom{00,}496.4
\\ 

\bottomrule
\end{tabular}
}\hfill\adjustbox{valign=c}{\includegraphics[width=.245\linewidth]{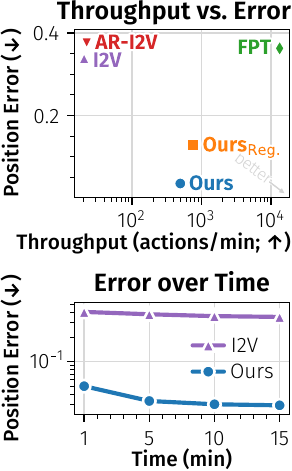}\vspace{-50em}}
    \caption{\textbf{Planning Billiard Shots through Future Exploration.} \textit{Left:} We compare the \textit{Accuracy} of landing a ball at a randomly selected goal position in a billiard simulation by unrolling potential futures starting from varying cue ball impulses. Under a fixed compute budget, our model surpasses dense world models from scratch using the same data. This is enabled by our methods' low \textit{Latency}, enabling us to sample a large number of potential futures. \textit{Right:} We visualize results w.r.t.\ final target error and show its evolution over planning time for our model and an I2V baseline.}
    \vspace{-1em}
    \label{tab:billiard_acc}
\end{table}

\subsection{Calibration}
\begin{figure}[t]
    \centering
    \includegraphics[width=0.7\linewidth]{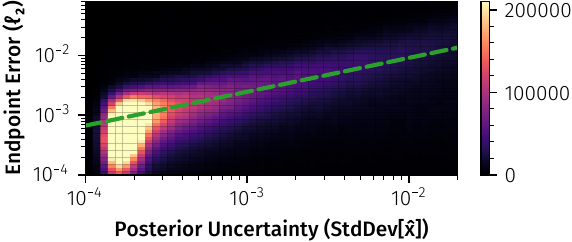}
    \caption{\textbf{Posterior Uncertainty vs.\ Error.} Starting around pixel-level ($\smash{\frac{1}{512}}$), our model's posterior uncertainty is well-correlated ({\color{matplotlibgreen}\textbf{green}} line) with true error.}
    \label{fig:uncertainty}
    \vspace{-1em}
\end{figure}

We explore the relation of our model's posterior uncertainty (as measured by standard deviation on the head's posterior) in \cref{fig:uncertainty}.
There is a large concentration around pixel-level error (error $< \smash{\frac{1}{512}}$); above that, the posterior uncertainty predicts the final error well (linear relation in log-log space).

\subsection{Ablations}
We ablate core architectural components, training for 400k steps on our open-set training data.
Performance metrics are calculated on {\ourdataname} following previous experiments.

\paragraph{Fast Reasoning Blocks}
We compare the inference speed of our efficient fused attention layers with that of a standard unfused attention layer, using self-attention for motion tokens and cross-attention for image tokens.
For a 32-timestep rollout (batch size 4, 16 trajectories), we achieve $\sim$2$\times$ faster sampling, enabling substantially more efficient exploration of the search space. This extends to $\sim$3.7$\times$ at batch size 1.

\paragraph{Posterior Parametrization}
Our model uses a point-wise FM head to represent the distribution over future motion. Furthermore, the input to the FM head is scaled using a cascade of exponentially separated value ranges, enabling the model to focus on different granularities of motion as needed. Removing the cascade results in a severe degradation in prediction quality as shown in \cref{tab:ablation_posterior}.

Alternatively, the distribution over future motion could be modeled with a Gaussian Mixture (GMM) Distribution head~\cite{tschannen2024givt} similar to the single-step Flow Poke Transformer~\cite{baumann2025whatif}. However, not only is the Gaussian Mixture constrained in what it can represent, leading to higher errors in \cref{tab:ablation_posterior}, the GMM head is also harder to train, converging more slowly as seen in \cref{tab:ablation_posterior}-right.

\begin{table}[t]
    \centering
    \adjustbox{max width=.6\linewidth}{\scalebox{.8}{
    \begin{tabular}{l@{\hskip .25em}c@{\hskip .35em}c}
        \toprule
        \textbf{Posterior Type} & \textbf{Scale Cascade} & \textbf{Best-5$\downarrow$} \\
        \midrule
        GMM~\cite{baumann2025whatif,tschannen2024givt} & n/a & 0.110 \\
        FM Head (Ours) & \xmark & \underline{0.033} \\
         FM Head (Ours) & \phantom{(Ours)}\cmark (Ours) & \textbf{0.029} \\
        \bottomrule
        \\
    \end{tabular}
    }}\hfill\adjustbox{valign=c}{\includegraphics[width=.395\linewidth]{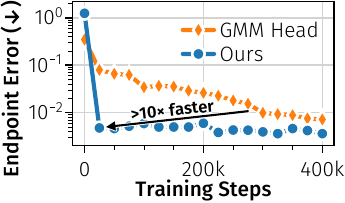}}
    \caption{\textbf{Posterior Parametrization Ablation.} {Substituting previously used GMM-based heads with flow matching heads leads to significant improvements in accuracy and increases convergence substantially. Adding our scale cascade improves accuracy further.}}
    \label{tab:ablation_posterior}
    \vspace{-1em}
\end{table}

\section{Conclusion}
\label{sec:conclusion}

To envision the many different futures of a scene in a stochastic open world, we have proposed an autoregressive diffusion model that can effectively explore the space of all potential trajectories step-by-step into the future. Our transformer-based model and a lightweight diffusion head model the multi-modal distribution of motion trajectories and allow for efficient training and inference, making our approach especially valuable under compute- and time-constrained settings.
The autoregressive approach also naturally lends itself to conditioning motion generation on user-provided initial motion, allowing for the exploration of the effects of actions under uncertainty about how the future will unfold.

To evaluate this setting, we presented \emph{OWM}, a benchmark for open-world motion prediction designed to test whether models can produce coherent, diverse trajectory distributions in realistic conditions. Across diverse domains -- from in-the-wild videos to controlled physical setups -- our method achieves accurate long-range predictions while dramatically reducing sampling cost, highlighting the advantage of directly modeling motion over future frame generation when accurate dynamics matter.
This efficiency of our model further facilitates rapid exploration of the space of possible actions and their outcomes to enable determining optimal action, such as how to select a billiard shot to take to achieve a specific outcome.
Taken together, our results highlight the value of a dynamics-centric representation for future reasoning. By focusing on how the world can move rather than how it should look, we provide an efficient, probabilistic mechanism for exploring possible futures -- one that can serve as a foundation for forecasting, planning, and interaction in complex real-world environments.

\paragraph{Limitations}
{
    Our main formulation assumes a static camera, which simplifies evaluation and improves interpretability of predictions, but limits applicability to scenes with ego-motion or dynamic viewpoints -- a setting that contemporary video generation baselines already handle.
    We explored a formulation that enables \emph{learning} from videos with dynamic cameras by compensating for it during preprocessing, but joint \emph{prediction} of ego and scene motion remains an important direction for future work.
    Additionally, our model relies on pseudo ground-truth trajectories from off-the-shelf trackers for training, inheriting their biases and failure modes.
}

\section*{Acknowledgments}
This project has been supported by a research grant from Netflix, the Horizon Europe project ELLIOT (GA No.\ 101214398), the project ``GeniusRobot'' (01IS24083) funded by the Federal Ministry of Research, Technology and Space (BMFTR), the BMWE ZIM-project (No.\ KK5785001LO4) ``conIDitional LoRA'', the German Federal Ministry for Economic Affairs and Energy within the project ``NXT GEN AI METHODS - Generative Methoden für Perzeption, Prädiktion und Planung'', and the bidt project KLIMA-MEMES. The authors gratefully acknowledge the Gauss Center for Supercomputing for providing compute through the NIC on JUWELS/JUPITER at JSC and the HPC resources supplied by the NHR@FAU Erlangen.
We thank Timy Phan, Nick Stracke, Kosta Derpanis, Kolja Bauer, Thomas Ressler-Antal, Frank Fundel, Enrico Shippole, Felix Krause, and Meimingwei Li for their helpful feedback and support, and Owen Vincent for continuous technical support.

\section*{Author Contributions}
SB and JW co-led the project.
SB conceived the initial idea (with BO), built the billiard prototype, and optimized the final model.
JW developed the final model and handled data processing and evaluation.
TM designed, curated, and implemented the OWM benchmark.
All authors contributed to writing.
MK and BO supervised the project and reviewed the manuscript.

{
    \small
    \bibliographystyle{ieeenat_fullname}
    \bibliography{main}
}

\clearpage

\clearpage
\setcounter{page}{1}
\maketitlesupplementary

\setcounter{figure}{0}
\setcounter{table}{0}
\setcounter{equation}{0}
\setcounter{section}{0}
\renewcommand\thefigure{\Alph{figure}}  
\renewcommand\thetable{\Alph{table}}  
\renewcommand\thesection{\Alph{section}}

\section{Additional Implementation Details}\label{sec:appendix_additional_details}
We provide more context on implementation details of our main model described in the paper.
Please also refer to the supplementary model code, which contains extensive further comments, for reference.

\subsection{Transformer Block}
We implement our transformer~\citep{vaswani2017attention,dosovitskiy2021an} blocks primarily following the standard Llama~\citep{touvron2023llama,touvron2023llama2}-style block architecture in a similar setup as \citet{baumann2025whatif}.
Specifically, we use pre-normalization with RMSNorm~\citep{zhang2019root}, omit bias terms in linear layers, and use rotary positional embeddings~\citep{su2021roformer} in an axial setup with scaled cosine similarity attention following \citet{crowson2024hourglass}.
Our feedforward network setup does not follow Llama's SwiGLU~\citep{shazeer2020glu} activation, but instead uses the more classical GELU~\cite{hendrycks2023gaussianerrorlinearunits}, while still retaining the omission of bias terms.
We observed that both choosing GELU with the typical $\tanh$ approximation as the activation and omitting the GLU-style~\cite{shazeer2020glu} gating leads to small speed improvements without significant decreases in quality.
Importantly, we implement a fully fused parallel transformer layer, where cross-attention and self-attention are combined into a single attention across both kinds of tokens, and projections are shared between the attention and feedforward network, as described in the main paper.

\subsection{Posterior Flow Matching Head}
Our flow matching posterior head follows similar high-level hyperparameters as \citet{li2024autoregressive}, with three layers of width 1024.
Unlike them, we use a standard flow matching~\cite{lipman2022flow} objective instead of the DDPM~\cite{ho2020denoising} formulation and perform substantial architectural changes to enable efficient sampling.
Each block is a standard pre-LayerNorm~\cite{ba2016layer} FFN block with GELU~\cite{hendrycks2023gaussianerrorlinearunits} activation.

\paragraph{Conditioning}
We implement conditioning such that every component can be cached. Typically, conditioning would be implemented with a local, per-layer MLP that projects a conditioning vector into channel scales, shifts, and, optionally, output gating coefficients. This causes a large number of extra kernel launches, which, as this head will perform tens to hundreds of forward passes per AR sampling step, would cause significant wall-clock overhead.
Instead, we precompute all scales and shifts centrally once.
Additionally, we factorize the conditioning on flow matching time $\tau$ and the conditioning on the parameters $\mathbf{z}_t\supi$ additively, such that the time conditioning can be precomputed offline once, and the parameter conditioning can be computed once per sampling loop, further reducing computational overhead. Conditioning inside each block is implemented via predicted scale and shift on the output of each pre-LayerNorm~\cite{ba2016layer}. We do not perform output gating.

\paragraph{Input Value ``Scale Cascade''}
For the posterior FM head, we use an input scale cascade to stabilize training when modeling motion.
Practically, this is implemented as a logarithmically spaced set of scale coefficients
\begin{equation}
    \mathbf{s} = \exp(\mathrm{linspace}(\log(0.1), \log(1e5), \mathrm{num}= 512)),
\end{equation}
with $\mathrm{linspace}(\mathrm{min}, \mathrm{max}, \mathrm{num})$ denoting the standard numpy/PyTorch operation, using which the features for the noisy input $x_\tau$ are computed component-wise as
\begin{equation}
    [\tanh(\mathbf{s} \cdot x_{\tau,0}) | \tanh(\mathbf{s} \cdot x_{\tau,1})]\mathbf{W}_{in}^\top,
\end{equation}
with $[\cdot|\cdot]$ denoting channelwise concatenation, and $\mathbf{W}_{in}$ being the output projection to the transformer's hidden dimension.

\paragraph{Sampling}
For sampling, we solve the ODE parametrized by the FM head using an Euler solver with uniform spacing of flow matching time $\tau$, matching our training setting of sampling from a uniform distribution $\tau \sim \mathcal{U}[0, 1]$. Unless specifically noted otherwise, we use 50 sampling steps.
During AR sampling, we simply sample one motion sample from the posterior, update the latest position of that trajectory, and then sample the next step defined by the AR factorization, while also conditioning on this new information.
This process can be started from partial motion information, initial motion hints (pokes), or no prior motion information.

\subsection{Hyperparameters}
\cref{tab:hyperparams} provides a comprehensive list of hyperparameters that describe our training setup and model configuration. We train the open-set motion model for 400k steps with a peak learning rate of $3 \times 10^{-5}$. We train with a linear learning rate warmup of 5000 steps, after which we apply a linear learning rate schedule.
The training setup for the Billiard simulation is similar, but trajectory positions are obtained from the Billiard physics engine~\cite{ebke2025pythonbilliards} and thus represent ground truth motion instead of tracker annotations. Further, we focus on longer-horizon prediction in the Billiard setup. We train the model to predict 50 timesteps, where each timesteps corresponds to a $\Delta t = 0.01\,s$ interval for 300k iterations.

\begin{table}[tbh]
    \centering
    \centering
\adjustbox{max width=\linewidth}{
\begin{tabular}{lccc} 
\toprule

\multirow{2}{*}[-3pt]{\centering\textbf{Parameter}}
& \multicolumn{3}{c}{\textbf{Value}} \\
\cmidrule(lr){2-4}
& \multicolumn{2}{c}{\textbf{Open-set}} & \textbf{Billiard} \\

\midrule

Dataset & Open-Set Videos & Open-Set Video 3$\rightarrow$2D  & Billiard Simulations \\
Number of clips & 10M & 1.5M & -- \\

\midrule

Tracker & TapNext~\cite{zholus2025tapnext} & V-DPM~\cite{sucar2026vdpm} & Ground-truth \\
Tracker position seeding & 1024 random positions & 16,641 grid positions & random ball starting positions \\
Flow scale & $[-1, 1]$ & $[-1, 1]$ & $[-1, 1]$  \\
Image size & $512 \times 512$ & $512 \times 512$ & $512 \times 512$ \\
Training track number & 16 & 16 & 16 \\
Training timesteps & 16 & 16 & 50 \\

\midrule

Batch size & 128 & 128 & 128 \\
Optimizer & AdamW~\cite{loshchilov2018decoupled} & AdamW~\cite{loshchilov2018decoupled} & AdamW~\cite{loshchilov2018decoupled} \\
Betas & (0.09, 0.99) & (0.09, 0.99) & (0.09, 0.99)\\
Peak learning rate & $3 \times 10^{-5}$ & $3 \times 10^{-5}$ & $3 \times 10^{-5}$ \\
Learning rate schedule & linear decay to $10^{-8}$ & linear decay to $10^{-8}$ & linear decay to $10^{-8}$ \\
Warm-up steps & 5k & 5k & 5k \\
Total steps & 400k & 400k & 300k \\
Precision & bfloat16 AMP & bfloat16 AMP & bfloat16 AMP \\
Total Parameters & 665M & 665M & 665M \\

\midrule

GPUs & 16 Nvidia H200 & 16 Nvidia H200 & 16 Nvidia H200s \\
Training Time & 20\,h & 20\,h & 20\,h \\

\midrule

Depth & 24 & 24 & 24 \\
Width & 1024 & 1024 & 1024 \\
Head dim & 128 & 128 & 128 \\
Normalization & RMSNorm & RMSNorm & RMSNorm\\
FFN expand factor & 4 & 4 & 4 \\
Activation & GELU & GELU & GELU \\
Positional Encoding & see \cref{sec:methodology} & see \cref{sec:methodology} & see \cref{sec:methodology} \\
Static scene conditioning & Adaptive Norm~\cite{huang2017arbitrary} & -- & -- \\
Denoiser width & 1024 & 1024 & 1024 \\
Denoiser depth & 3 & 3 & 3 \\

\bottomrule
\end{tabular}
}

    \caption{Hyperparameters of our main models and training setup.}
    \label{tab:hyperparams}
\end{table}

\subsection{Training Data}

We use three sources of training data for our models.

\paragraph{Open-set Video Data}
To train our model for open-world motion generation, we source diverse videos from the internet, while ensuring no overlap with our evaluation data. We then apply an off-the-shelf tracker~\cite{zholus2025tapnext} to obtain pseudo ground-truth annotations. For training, we center crop images to square resolution, cropping in both axis slightly to avoid border points for which the tracker commonly fails. We then resize frames to a uniform $512 \times 512$ resolution.

\paragraph{Reprojected 3D Data}
Large-scale open-set videos typically suffer from ego camera motion, limiting the interpretability of trajectories. We aim to train a motion model, predicting interpretable static camera trajectories on unconstrained video data for scalability.
We thus apply V-DPM~\cite{sucar2026vdpm} a 3D tracker that also estimates camera motion to open-set videos. Then, we reproject tracks into the first camera view, resulting in stabilized trajectories without camera motion interference. We apply the same center crop and resize.

\paragraph{Billiard Data}
Training data for the Billiard simulation is obtained using a billiard physics simulation~\cite{ebke2025pythonbilliards}. Ball positions and velocity are sampled randomly while ensuring balls do not overlap with other balls or the border. The physics engine produces future positions of balls, which are used as tracks to train the model.

\subsection{Benchmark Construction}
To create the \ourdataname dataset, we source 95 permissively licensed videos from Pexels\footnote{\url{https://www.pexels.com/}} that have been verified to have a static camera and cover a large variety of different kinds of motion from different kinds of entities (\emph{e.g.}, people, vehicles, animals, objects, ...). We prioritize structured or kinematically constrained dynamics (\emph{e.g.}, articulated bodies, rigid object movement) and avoid stochastic or disconnected movement (\emph{e.g.}, excessive background movement, excessively unconstrained motion). We further manually annotate a start frame and select points of interest on moving objects. Ground truth trajectories are obtained with TAPNext \cite{zholus2025tapnext} and the tracking quality is manually verified.

We complement our dataset with samples from existing solid mechanics benchmarks with known high complexity. For this purpose, we obtain 97 samples from Physics-IQ~\cite{motamed2025generative} (subset ``solid mechanics'') and 134 samples from Physion~\cite{bear2021physion} (excluding the ``Drape'' subset because of it's focus on soft-body collisions). We manually verify the correctness of motion in the Physion subset, as we observed some examples with unrealistic physical simulation. We, again, manually select starting frames and query points, and verify the correctness of motion annotations for all the additional samples.

\subsection{Metrics}

\paragraph{Open-World Motion Prediction}
For the open-world and physical motion prediction benchmark, we rely on a simple MSE objective between the ground truth trajectory points $\mathbf{p}_{gt}$ and the predicted trajectory $\mathbf{p}_{pred}$ by evaluated methods, where $\mathbf{p}$ is a sequence of $T$ 2D points $\mathbf{p} \in \mathbb{R}^{T \times 2}$. The ground truth is obtained by applying TapNext~\cite{zholus2025tapnext} to the full original video. As in a given initial configuration, multiple outcomes could be reasonable, we give each method the chance to produce an ensemble of predictions, whereby the ensemble size is $N_{ens} = 5$ for the \textbf{Best-of-5} setting and $N_{ens}$ depends on the throughput of each method in the \textbf{Best-in-5min} setting.
Throughput is calculated using best effort, meaning we utilize optimized implementations and lower-precision calculations when possible.

\paragraph{Billiard Planning}
We calculate throughput similarly under optimized settings. To calculate the planning accuracy, we use the best action found during rollouts using the principle from \cref{eq:expected_reward}. Then, we perform rollouts of the true Billiard simulation using the found action as the initial motion, while all balls except the action ball are initialized as stationary. A selected action is counted as correct if the target ball at least touches or covers the predefined goal position within the allocated time frame. If not the selected action is counted as incorrect. The accuracy is then calculated by dividing the number of correct actions $N_{correct}$ by the total number of trials $N_{total}$.

\subsection{Baselines}

\paragraph{Open-World Baselines}
For the open-world and physics evaluation, we compare against five state-of-the-art video generation models: MAGI-1 4.5B~\cite{teng2025magi}, Wan2.2 I2V-A14B~\cite{wan2025}, CogVideo-X 1.5 5B-I2V~\cite{yang2024cogvideox}, SkyReels V2 DF 1.3B 540P~\cite{chen2025skyreelsv2infinitelengthfilmgenerative}, and Stable Video Diffusion 1.1 (SVD)~\cite{blattmann2023stable}.
We utilize the implementation provided in the diffusers~\cite{von-platen2022diffusers, wolf2020huggingfaces} library for Wan, CogVideo-X, SkyReels, and SVD. For MAGI, no diffusers implementation is available as of the writing of the paper, therefore we instead adopt the official repository and checkpoint and use the provided 4.5B distill+quant variant.
All models except MAGI are run in I2V mode. Thus, they receive the last known image as conditioning and are tasked to simulate the video rollout. As multiple continuations are possible, we sample the Best-out-of-5 and Best-in-5min motion, respectively, giving the models the chance to explore multiple possible outcomes under uncertainty.
For MAGI-1, we run the model in video-2-video mode and provide frames preceding the last known frame as hint conditioning.
We subsequently apply TapNext~\cite{zholus2025tapnext} tracking to generated videos to obtain predicted trajectories, which we use to compute metrics.

\paragraph{Billiard Baselines}
We compare billiard action search performance against four video generation baselines and two trajectory prediction baselines, which we implement and train from scratch to ensure fair comparison.
We match the training setup as closely as possible to the setup for our model.

Video generation models are implemented as image-conditioned spatio-temporal Diffusion Transformers~\cite{peebles2023scalable}. For efficiency, we utilize latent diffusion~\cite{rombach2022highresolution} and perform diffusion in the latent space of the pretrained VAE from Stable Diffusion-XL~\cite{podell2023sdxl}. Image-conditioning is achieved by cross-attending to the VAE-produced tokens of the start image.
We train four variants of video diffusion models, differing along two axes to cover a variety of previous approaches.
Our video diffusion models either use auto-regressive generation or full sequence diffusion. In the former setting, the image conditioning is auto-regressively updated to include the prior $N_{hist}$ images. The auto-regressive video generation model then generates the single next frame, conditioned on the history of previous images.
The full sequence diffusion approach, on the other hand, is conditioned solely on the initial image and generates the full video from a single noise sample $x_1 \in \mathbb{R}^{T \times H \times W \times C}$.
The models further differ in how they are informed about motion prompts. The \textit{Images to Video} variants receive an additional second conditioning image to which they cross-attend. Note that this is natively supported by AR video generation models, while full sequence diffusion requires modification. Therefore, these models can infer the initial motion from visual cues.
The \textit{poke-cond.}\ models receive the instantaneous flow as an additional conditioning similar to our method. The flow and positions are first embedded using Fourier Embeddings and then passed through a small-scale MLP before being pooled into a fixed-size vector with a linear layer for multiple trajectories. The model is then conditioned on the flow embedding using Adaptive Layer Normalization~\cite{ba2016layer,huang2017arbitrary}.
We use L-sized DiT backbones~\cite{peebles2023scalable} for our experiments and train the video diffusion models until convergence.

For the full trajectory diffusion baseline, we ensure a fair comparison by reusing our motion models' backbone, but replacing the auto-regressive point-wise diffusion head with a DiT~\cite{peebles2023scalable}. The training setup and motion model hyperparameters are consistent with our standard setup; however, we ensure that the model always receives only the first step flow.

For the FPT~\cite{baumann2025whatif} baseline, we utilize the official implementation and train the model for 2 million steps. Note that all other models predict step-wise motion, while FPT samples future positions in a single step.
We align the horizon of the FPT baseline with that of the step-wise models and predict the final position of the balls at the end of the prediction window.

\section{Additional Ablations}\label{sec:appendix_additional_ablations}
In the following, we elaborate further on design choices in our implementation.

\subsection{Number of Function Evaluations}
We test the impact of using more evaluations of the denoising flow matching head on the endpoint error (EPE) in the Billiard setting.
Results in \cref{tab:nfe_epe} show that our approach yields lower endpoint error with more function evaluations.
Beyond 10 function evaluations, the benefits begin to diminish. Therefore, for our main evaluations in \cref{sec:experiments} we use 50 evaluations to balance quality and speed.

\begin{table}[tbh]
    \centering
    \adjustbox{max width=\linewidth}{
        \scalebox{.8}{
            \centering
\begin{tabular}{rc} 
\toprule

\textbf{NFEs} & \textbf{Mean-best-of-5-EPE} \\

\midrule

1
& 0.00361
\\

5
& 0.00143
\\

10
& 0.00140
\\

25
& \underline{0.00139}
\\

50
& \textbf{0.00138}
\\

\bottomrule
\end{tabular}

        }
    }
    \caption{\textbf{Inference Time Scaling:} Our approach achieves lower End-Point-Error in the Billiard simulation with more function evaluations of the diffusion head.}
    \label{tab:nfe_epe}
\end{table}

\subsection{Trajectory ID Embedding}
As outlined in \cref{sec:methodology} we draw random, (nearly) orthogonal trajectory embeddings $\text{id}_{\text{traj}}^{(i)} \sim \mathcal{U}(\mathbb{S}^{d-1})$ to indicate trajectory correspondence to the model. Other, more common approaches would be to use no explicit embedding and instead only rely on positional embeddings, or to use a learnable trajectory embedding with a fixed-size codebook.

We compare these options in \cref{tab:traj_emb} on the Billiard simulation data.
We find that our randomized embeddings outperform both learnable embeddings (likely attributable to a reduction in the likelihood of the model learning position-related biases) and the setting with no extra embeddings. Importantly, unlike learnable embeddings, the model is capable of zero-shot trajectory number extrapolation from 16 (the number observed during training) to larger and smaller numbers, with minimal performance degradation.

\begin{table}[tb]
    \centering
    \adjustbox{max width=\linewidth}{
        \scalebox{.8}{
        \centering
\begin{tabular}{lrc} 
\toprule

\textbf{Traj. Emb.} & \textbf{Num. Traj.} & \textbf{Mean-best-of-5-EPE}\\

\midrule

\multirow{3}{*}{No Emb.}
& 8
& 0.00116\\

& 16
& 0.00150\\

& 24
& 0.00277 \\

\arrayrulecolor{gray!50!white}
\midrule
\arrayrulecolor{black}

\multirow{3}{*}{Learnable}
& 8
& 0.00112 \\

& 16
& 0.00149 \\

& 24
& not possible \\

\arrayrulecolor{gray!50!white}
\midrule
\arrayrulecolor{black}

\multirow{3}{*}{Ours}
& 8
& \textbf{0.00108} \\

& 16
& \textbf{0.00141} \\

& 24
& \textbf{0.00263} \\

\bottomrule
\end{tabular}

        }
    }
    \caption{\textbf{Trajectory ID Embedding:} Our trajectory ID embeddings provide {lower end-point-error} in billiard simulations and {enable zero-shot generalization} to \textit{both} increased and reduced number of trajectories.}
    \label{tab:traj_emb}
\end{table}

\subsection{Multi-Step Reasoning}

\begin{table}[tbh]
    \centering
    \adjustbox{max width=\linewidth}{
        \scalebox{.8}{
        \centering
\begin{tabular}{llcc} 
\toprule

$\mathbf{T}$ & $\boldsymbol{\Delta} \mathbf t$ & \textbf{Num. Steps} & \textbf{Mean-best-of-5-EPE}\\

\midrule

0.5 & 0.01
& 50
& \textbf{0.00141}
\\

0.5 & 0.05
& 10
& \underline{0.00999}
\\

0.5 & 0.5
& \phantom{0}1
& 0.02823
\\

\bottomrule
\end{tabular}

        }
    }
    \caption{\textbf{Reasoning in multiple steps.} We compare predicting 0.5\,s into the future using models trained with different step sizes. Our standard method integrates 50 steps, while the other models perform fewer steps. Therefore, these models require fewer auto-regressive steps, yet have to model more of the dynamics internally.}
    \label{tab:multi_step}
\end{table}

Our approach predicts the motion over a short time horizon $\Delta t$ in one step and then auto-regressively samples movement to predict motion over the entire time horizon $T$, thus factorizing the full motion prediction over $\Delta T$ into a sequence of small-step predictions.
In theory, a motion model with infinite capacity should be able to predict the final position of all scene elements in a single step by internally accounting for all potential interactions.
However, we argue that predicting step-wise motion is a substantially more practically viable task when not assuming abundant model capacity.
We investigate this assumption by comparing model variants in the Billiard setting.

We compare our standard model, predicting $\Delta t = 0.01\,s$ into the future per step, against variants predicting over a larger $\Delta t$. We perform a 0.5\,s rollout (making the largest-step model perform predictions over the full horizon $T$ in a single step, as in \cite{baumann2025whatif}) and evaluate the end-point-error of each model.
The results in \cref{tab:multi_step} show that multi-step motion prediction improves modeling performance, with overall improved performance for smaller step intervals.
The single-step model performs significantly worse than both multi-step variants, mirroring the planning results in \cref{tab:billiard_acc}. We attribute this failure to the complexity of internally modeling and accounting for all interactions in a large $\Delta t$ timeframe.

\subsection{Classic Trajectory Forecasting Setting}
We explore our approach's efficacy in classic trajectory forecasting settings in a \emph{zero-shot} setting.
We compare on the canonical ETH-UCY~\cite{pellegrini2009eth, lerner2007ucy} benchmark following the setting of Trajectron++~\cite{salzmann2020trajectron++}.
All baselines are trained exclusively on in-domain data, while we apply our model zero-shot.
The baselines directly operate on tracked abstract agents in a 2D top-down view space (obtained from the camera-space tracks via projection), while we operate directly on the original images, as our model uses that as the input.
Since the given homographies are not accurate for reprojecting back into the camera space, we manually annotate correspondences and fit homographies, obtaining the equivalent tracks in camera space, which serve as input and output space for our model.
ETH generally annotates people's heads, while UCY seems to rely on people's feet. This does not matter in a top-down view, as the head will typically be in the same 2D position as the feet, but it matters in camera space. We annotate homographies to follow the ETH convention.
Metrics are computed in the original 2D top-down space directly following the baselines.

We show results in \cref{tab:human_trajectory_eth_ucy}.
Despite not being trained for this setting, our method achieves competitive results with canonical task-specific baselines.
This demonstrates that our much more generic approach can still perform well even in specific settings.
With additional finetuning on sufficiently large-scale in-domain data, results should further improve significantly.

\begin{table*}[ht]
    \centering
    \adjustbox{max width=\linewidth}{
    \begin{tabular}{lcccccccc}
    \toprule
    \multirow{2}{*}[-3pt]{Method} & \multicolumn{2}{c}{ETH} & \multicolumn{2}{c}{Hotel} & \multicolumn{2}{c}{Zara01} & \multicolumn{2}{c}{Zara02} \\
    \cmidrule(lr){2-3}\cmidrule(lr){4-5}\cmidrule(lr){6-7}\cmidrule(lr){8-9}
     & Deterministic & Best-of-20 & Deterministic & Best-of-20 & Deterministic & Best-of-20 & Deterministic & Best-of-20 \\
    \midrule
    SocialLSTM~\cite{alahi2016social}          & 1.09/2.35 & --        & 0.79/1.76 & --        & \underline{0.47}/\underline{1.00} & --        & \underline{0.56}/\underline{1.17} & -- \\
    SocialGAN~\cite{gupta2018social}           & --        & 0.81/1.52 & --        & 0.72/1.61 & --        & \underline{0.34}/\underline{0.69} & --        & \underline{0.34}/\underline{0.69} \\
    Trajectron~\cite{ivanovic2019trajectron}          & --        & 0.59/1.14 & --        & 0.35/0.66 & --        & 0.43/0.83 & --        & 0.43/0.83 \\
    Trajectron++~\cite{salzmann2020trajectron++}        & \textbf{0.71}/\underline{1.66} & \underline{0.39}/\underline{0.83} & \textbf{0.22}/\textbf{0.46} & \textbf{0.12}/\textbf{0.19} & \textbf{0.39}/\textbf{0.77} & \textbf{0.15}/\textbf{0.33} & \textbf{0.23}/\textbf{0.59} & \textbf{0.11}/\textbf{0.25} \\
    \modelname (ours, zero-shot)    & \underline{0.81}/\textbf{1.50} & \textbf{0.31}/\textbf{0.80} & \underline{0.30}/\underline{0.54} & \underline{0.17}/\underline{0.30} & 0.79/1.75 & 0.53/1.21 & 0.58/1.30 & 0.40/0.91 \\
    \bottomrule
    \end{tabular}
    }
    \caption{\textbf{Zero-shot Comparison with Closed-Domain Trajectory Forecasting on ETH-UCY~\cite{pellegrini2009eth,lerner2007ucy}.} All numbers (except ours) are sourced from Trajectron++~\cite{salzmann2020trajectron++}. Note that some sequences from UCY are missing due to missing RGB videos. Values are (following Trajectron++) (min-)\{ADE/FDE\}, ``--'' means not reported by Trajectron++. In the ``deterministic'' setting, we sample from our model once with fixed seed.}
    \label{tab:human_trajectory_eth_ucy}
\end{table*}

\subsection{OWM Breakdown}
We report metrics for subsets of OWM focusing on specific kinds of motion in \cref{tab:owm_subset_metrics}. Our model is competitive with substantially larger video baselines for all subsets, including intricate multi-agent interactions. In the time constraint setting our method achieves the best results across all subsets as it's fast inference allows to explore a much larger variety of potential futures.

\begin{table*}[t]
    \centering
    \adjustbox{max width=\linewidth}{
    \begin{tabular}{lc@{}cc@{}cc@{}cc@{}cc@{}cc@{}c|c@{}ccc}
        \toprule
        & \multicolumn{4}{c}{Rigidity} & \multicolumn{4}{c}{Number of Agents} & \multicolumn{4}{c|}{Agents with Free Will} \\
        \cmidrule(lr){2-5} \cmidrule(lr){6-9}\cmidrule(lr){10-13}  
        \multirow{2}{*}[-3pt]{Method} & \multicolumn{2}{c}{Rigid} & \multicolumn{2}{c}{Non-rigid} & \multicolumn{2}{c}{Single-Agent} & \multicolumn{2}{c}{Multi-Agent} & \multicolumn{2}{c}{w/ Free Will} & \multicolumn{2}{c}{w/o Free Will} & \multicolumn{2}{|c}{Avg.\ Rank} & \multirow{2}{*}[-3pt]{\shortstack{Throughput\\{\footnotesize \textsc{samples/min$\uparrow$}}}} & \multirow{2}{*}[-3pt]{Params$\downarrow$} \\
        \cmidrule(lr){2-3} \cmidrule(lr){4-5} \cmidrule(lr){6-7} \cmidrule(lr){8-9} \cmidrule(lr){10-11} \cmidrule(lr){12-13} \cmidrule(lr){14-15}
        & {\footnotesize \textsc{N=5$\downarrow$}} & {\footnotesize \textsc{T=5min$\downarrow$}} & {\footnotesize \textsc{N=5$\downarrow$}} & {\footnotesize \textsc{T=5min$\downarrow$}} & {\footnotesize \textsc{N=5$\downarrow$}} & {\footnotesize \textsc{T=5min$\downarrow$}} & {\footnotesize \textsc{N=5$\downarrow$}} & {\footnotesize \textsc{T=5min$\downarrow$}} & {\footnotesize \textsc{N=5$\downarrow$}} & {\footnotesize \textsc{T=5min$\downarrow$}} & {\footnotesize \textsc{N=5$\downarrow$}} & {\footnotesize \textsc{T=5min$\downarrow$}} & {\footnotesize \textsc{N=5$\downarrow$}} & {\footnotesize \textsc{T=5min$\downarrow$}} \\
        \midrule
        MAGI-1 & \vtwo{0.032} & \vthree{0.058} & \vtwo{0.039} & \vthree{0.069} & \vone{0.020} & \vtwo{0.044} & \vthree{0.048} & 0.080 & \vthree{0.040} & \vthree{0.066} & \vone{0.030} & \vthree{0.065} & \vone{2.00} & 3.00 & 0.303 & 4.5B \\
        Wan2.2~\cite{wan2025} & \vthree{0.042} & DNF & \vone{0.038} & DNF & 0.039 & DNF & \vone{0.039} & DNF & \vone{0.036} & DNF & 0.045 & DNF & \vthree{2.33} & DNF & 0.141 & 14B \\
        CogVideo-X 1.5~\cite{yang2024cogvideox} & 0.051 & DNF & 0.051 & DNF & 0.041 & DNF & 0.052 & DNF & 0.049 & DNF & 0.054 & DNF & 4.50 & DNF & 0.051 & 5B \\
        SkyReels V2~\cite{chen2025skyreelsv2infinitelengthfilmgenerative} & 0.061 & 0.071 & 0.056 & \vtwo{0.066} & 0.048 & 0.056 & 0.064 & \vtwo{0.075} & 0.054 & \vtwo{0.063} & 0.065 & 0.076 & 5.50 & 3.00 & \vthree{0.304} & \vtwo{1.3B} \\
        SVD 1.1~\cite{blattmann2023stable} & 0.048 & \vtwo{0.055} & 0.057 & 0.073 & \vthree{0.037} & \vthree{0.053} & 0.065 & \vthree{0.077} & 0.060 & 0.069 & \vthree{0.042} & \vtwo{0.064} & 4.66 & 3.00 & \vtwo{0.714} & \vthree{1.5B} \\
        \modelname (Ours) & \vone{0.031} & \vone{0.007} & \vtwo{0.039} & \vone{0.016} & \vtwo{0.036} & \vone{0.008} & \vtwo{0.044} & \vone{0.017} & \vtwo{0.037} & \vone{0.014} & \vtwo{0.044} & \vone{0.011} & \vone{2.00} & \vone{1.00} & \vone{2200} & \vone{0.6B} \\
        \bottomrule
    \end{tabular}
    }
    \caption{\textbf{OWM Subset-wise Metrics.} Breakdown of \cref{tab:phys_correctness} results. While orders of magnitude faster, our method is consistently competitive with state-of-the-art video models across not only the overall benchmark, but also when split across multiple properties (rigidity of motion, number of agents, presence of agents with free will).}
    \label{tab:owm_subset_metrics}
\end{table*}

\section{Additional Qualitative Samples}\label{sec:appendidix_additional_qualitative}

\begin{figure*}[h]
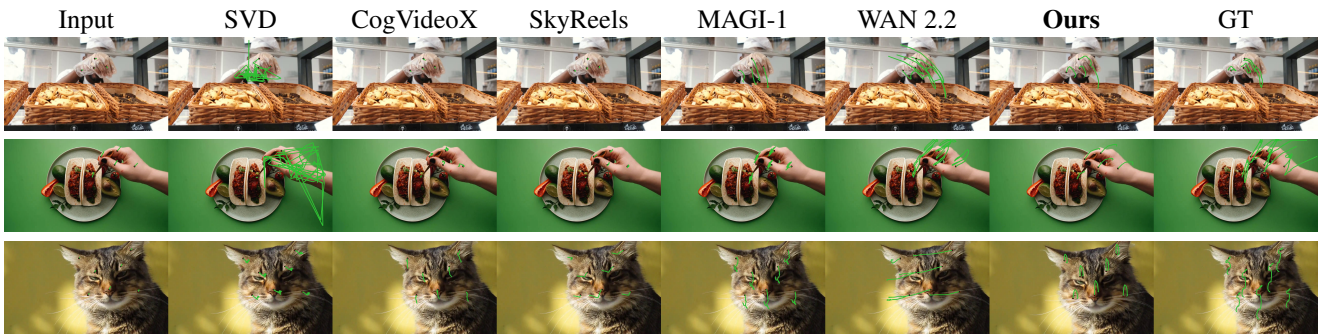

\centering
\newlength\q
\setlength\q{\dimexpr .125\linewidth -2\tabcolsep}
\begin{tabular}{>{\centering\arraybackslash}p{\q}>{\centering\arraybackslash}p{\q}>{\centering\arraybackslash}p{\q}>{\centering\arraybackslash}p{\q}>{\centering\arraybackslash}p{\q}>{\centering\arraybackslash}p{\q}>{\centering\arraybackslash}p{\q}>{\centering\arraybackslash}p{\q} @{}}
    Input & SVD & CogVideoX & SkyReels & MAGI-1 & WAN 2.2 & \textbf{Ours} & GT \\
    \multicolumn{8}{@{}c@{}}{\includegraphics[width=\textwidth]{"fig/qual/comparison/owm/row_1.jpg"}} \\
    \multicolumn{8}{@{}c@{}}{\includegraphics[width=\textwidth]{"fig/qual/comparison/owm/row_2.jpg"}} \\
    \multicolumn{8}{@{}c@{}}{\includegraphics[width=\textwidth]{"fig/qual/comparison/owm/row_3.jpg"}} \\
\end{tabular}
\caption{\textbf{Qualitative comparison on OWM:} Our model produces motion samples that are qualitatively on par with much larger models such as WAN2.2 and MAGI-1.}
\label{fig:comparison_omw}
\end{figure*}

\begin{figure*}[h]
\centering
\setlength\q{\dimexpr .125\linewidth -2\tabcolsep}
\begin{tabular}{>{\centering\arraybackslash}p{\q}>{\centering\arraybackslash}p{\q}>{\centering\arraybackslash}p{\q}>{\centering\arraybackslash}p{\q}>{\centering\arraybackslash}p{\q}>{\centering\arraybackslash}p{\q}>{\centering\arraybackslash}p{\q}>{\centering\arraybackslash}p{\q} @{}}
    Input & SVD & CogVideoX & SkyReels & MAGI-1 & WAN 2.2 & \textbf{Ours} & GT \\
    \multicolumn{8}{@{}c@{}}{\includegraphics[width=\textwidth]{"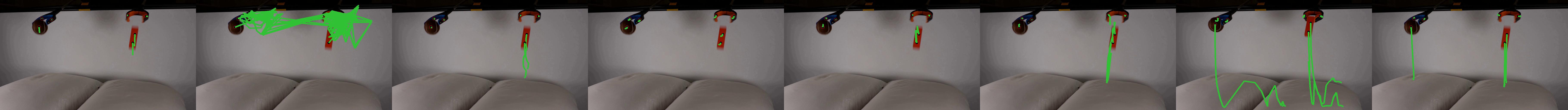"}} \\
    \multicolumn{8}{@{}c@{}}{\includegraphics[width=\textwidth]{"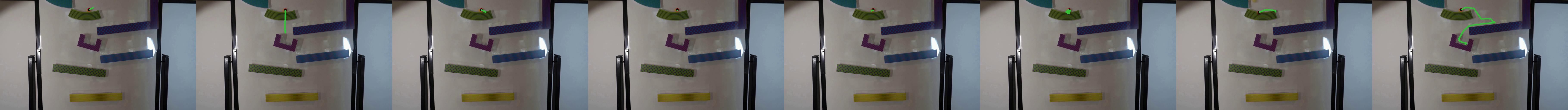"}} \\
    \multicolumn{8}{@{}c@{}}{\includegraphics[width=\textwidth]{"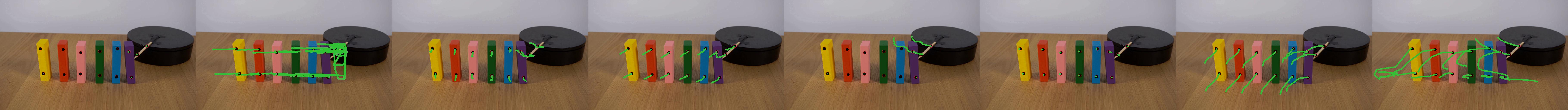"}} \\
\end{tabular}
\caption{\textbf{Qualitative comparison on Physics-IQ:} While video generation models fail to capture the complexity of object interactions and predict simplified or no motion, our approach captures realistic physical interactions.}
\label{fig:comparison_physics_iq}
\end{figure*}

\begin{figure*}[h]
\centering
\setlength\q{\dimexpr .125\linewidth -2\tabcolsep}
\begin{tabular}{>{\centering\arraybackslash}p{\q}>{\centering\arraybackslash}p{\q}>{\centering\arraybackslash}p{\q}>{\centering\arraybackslash}p{\q}>{\centering\arraybackslash}p{\q}>{\centering\arraybackslash}p{\q}>{\centering\arraybackslash}p{\q}>{\centering\arraybackslash}p{\q} @{}}
    Input & SVD & CogVideoX & SkyReels & MAGI-1 & WAN 2.2 & \textbf{Ours} & GT \\
    \multicolumn{8}{@{}c@{}}{\includegraphics[width=\textwidth]{"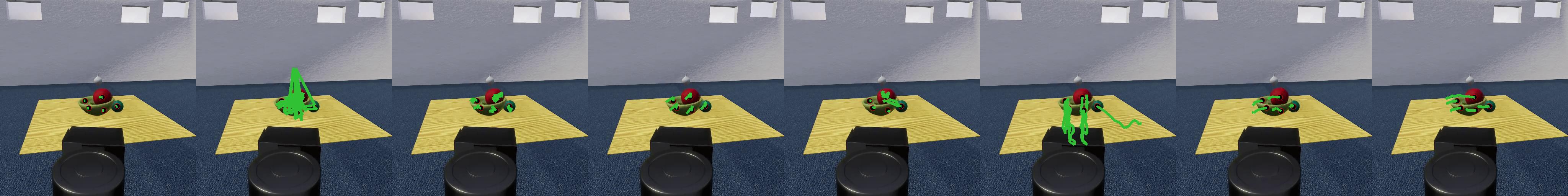"}} \\
    \multicolumn{8}{@{}c@{}}{\includegraphics[width=\textwidth]{"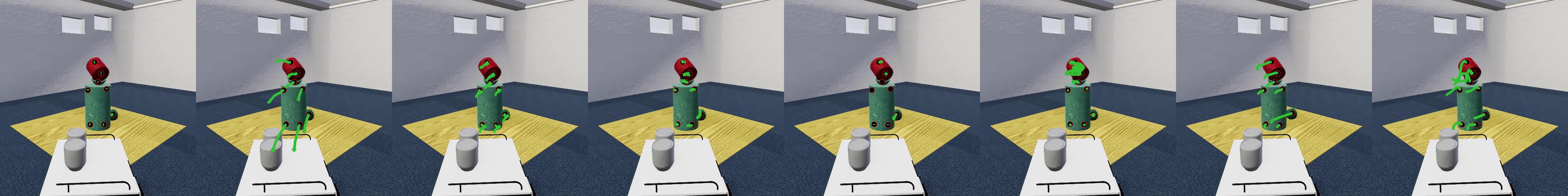"}} \\
    \multicolumn{8}{@{}c@{}}{\includegraphics[width=\textwidth]{"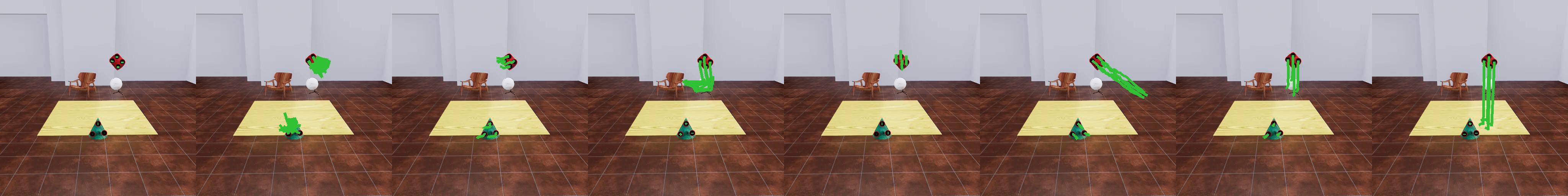"}} \\
\end{tabular}
\caption{\textbf{Qualitative comparison on Physion:} For simplified rigid body settings in Physion, video generation models hallucinate overly complex motion, while our approach is able to capture physical dynamics.}
\label{fig:comparison_physion}
\end{figure*}

\paragraph{Benchmark samples}  We provide qualitative samples from the \ourdataname benchmark (\cref{fig:comparison_omw}), Physics-IQ subset (\cref{fig:comparison_physics_iq}), and Physion subset (\cref{fig:comparison_physion}) with the \textbf{Best-out-of-5} motion annotation for our approach and all baseline methods.

Qualitatively, our approach predicts motion that is on par with state-of-the-art video generation approaches in open-world settings, found in the \ourdataname benchmark.
Comparing on Physics-IQ~\cite{motamed2025generative}, video generation approaches tend to predict overly simplified, physically implausible trajectories, whereas our method is able to capture the complexity of real-world physical interactions.
For Physion~\cite{bear2021physion}, state-of-the-art video generation models hallucinate overly complex motion, whereas our approach is able to capture the rigid body physics of the benchmark setting. Therefore, our approach is able to balance complexity and simplicity better than previous approaches making it applicable to a wider range of inference contexts.

\begin{figure*}[t]
    \newcommand{\imgwidth}{0.175\linewidth}
    \newcommand{\insertimage}[1]{\includegraphics[width=\imgwidth]{fig/qual/informed/#1_informed.png}}
    \newcommand{\imagerow}[2]{\insertimage{#1}&\insertimage{#2}}
    \centering
    \begin{tabular}{cccc}
        \imagerow{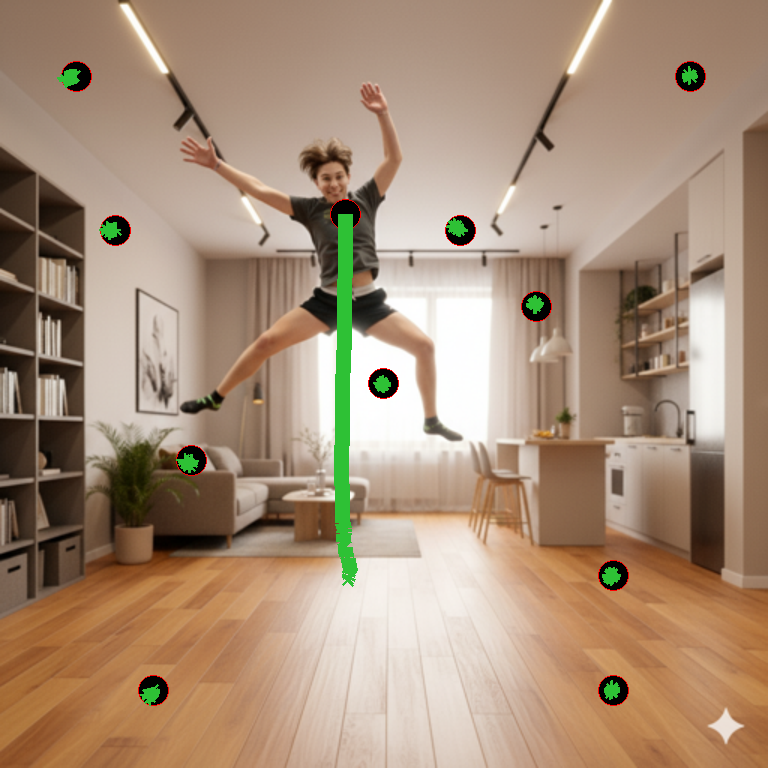}{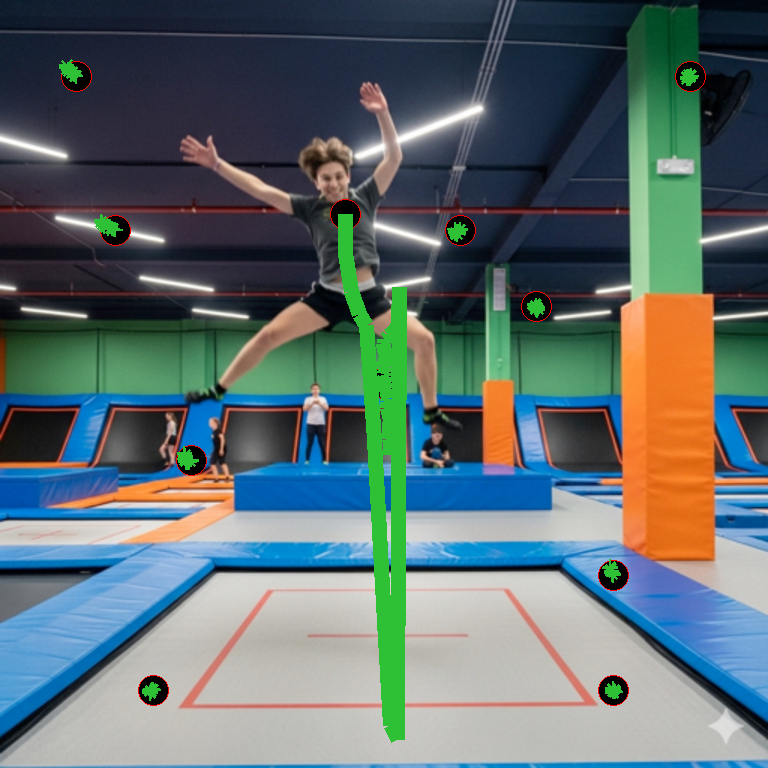} & \imagerow{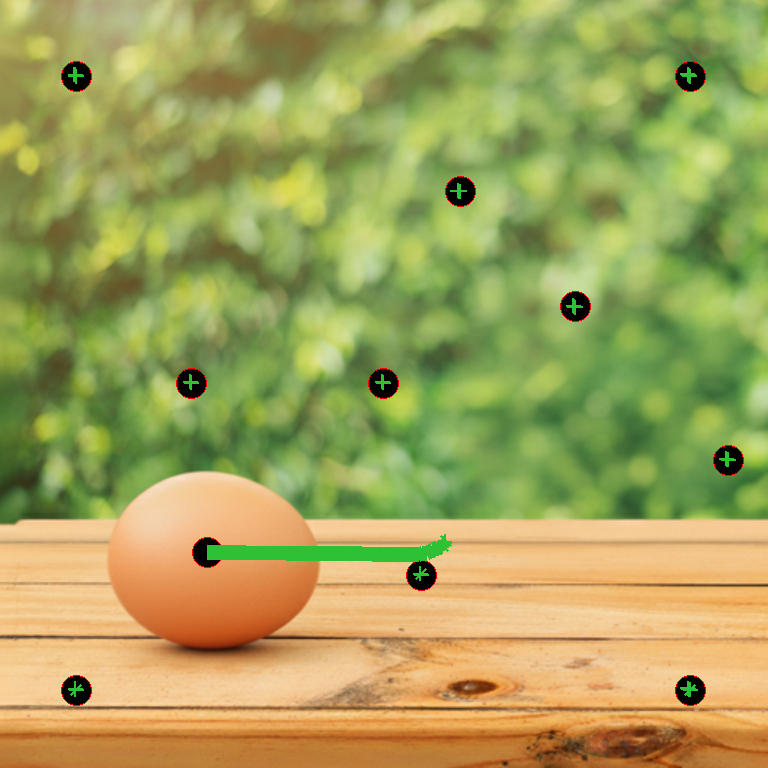}{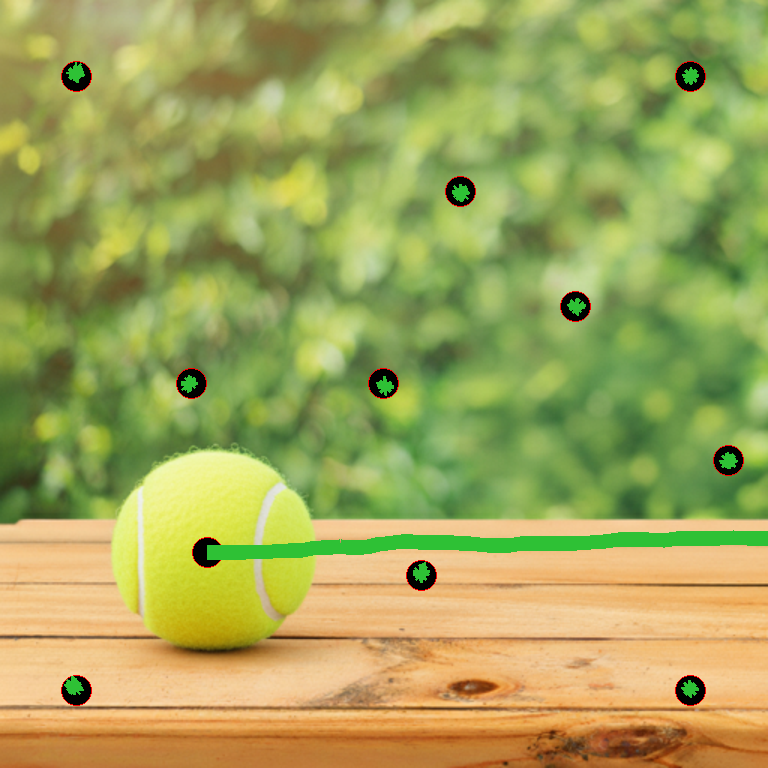}\\
    \end{tabular}
    \caption{\textbf{Context informed samples:} The above samples show our model's ability to take appearance information into account when predicting motion. In both comparisons, images were sampled with the same initial poke. Images where edited using nano banana for high similarity in appearance.}
    \label{fig:informed_qual}
\end{figure*}

\paragraph{Open-set samples}
We provide samples for a variety of open-set conditioning images, sourced from the internet. 
\cref{fig:informed_qual} shows that our approach predicts motion informed by the context provided through the starting image. We provide two examples where we edit the image using nano banana and use the same motion hint. The qualitative samples show that for a person on a trampoline, more {bouncy} motion is predicted compared to a person jumping on a wooden floor. Further, a ball rolling across a table has a more straight trajectory compared to an egg rolling across the same table.

\begin{figure*}[t]
    \newcommand{\imgwidth}{0.175\linewidth}
    \newcommand{\insertimage}[1]{\includegraphics[width=\imgwidth]{fig/qual/poked/#1_poked.png}}
    \newcommand{\imagerow}[4]{\insertimage{#1}&\insertimage{#2}&\insertimage{#3}&\insertimage{#4}}
    \centering
    \begin{tabular}{cccc}
        \imagerow{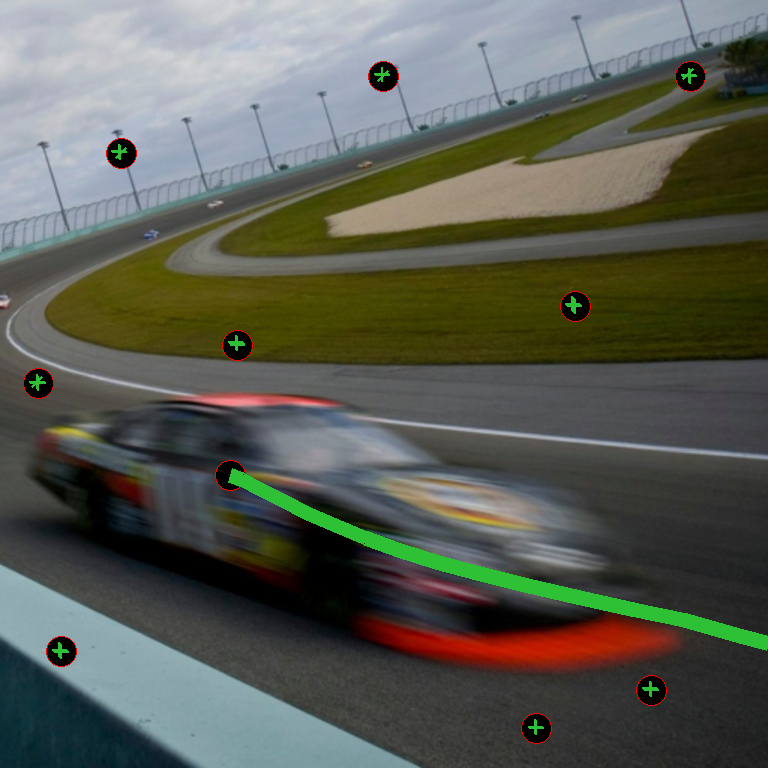}{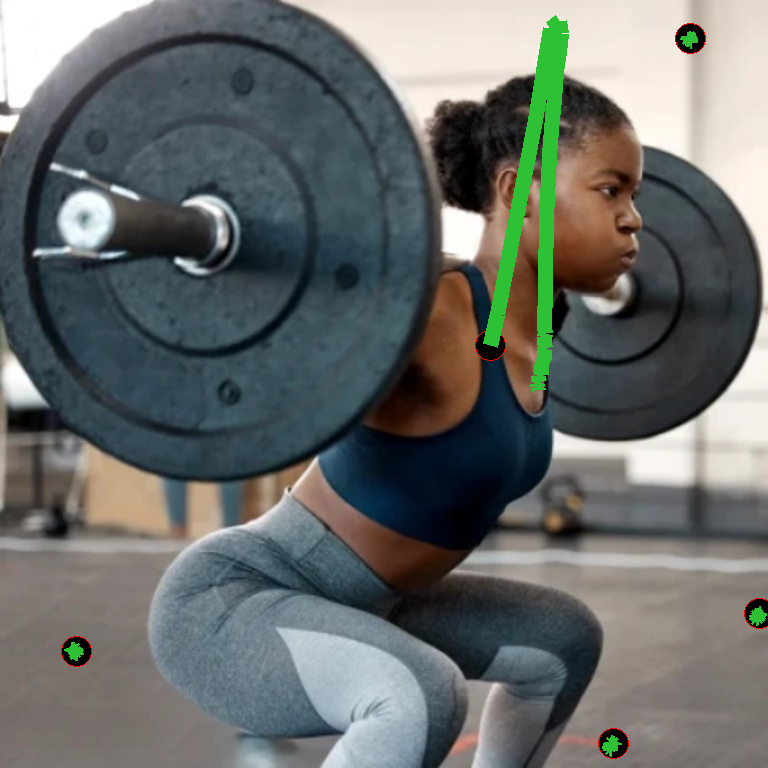}{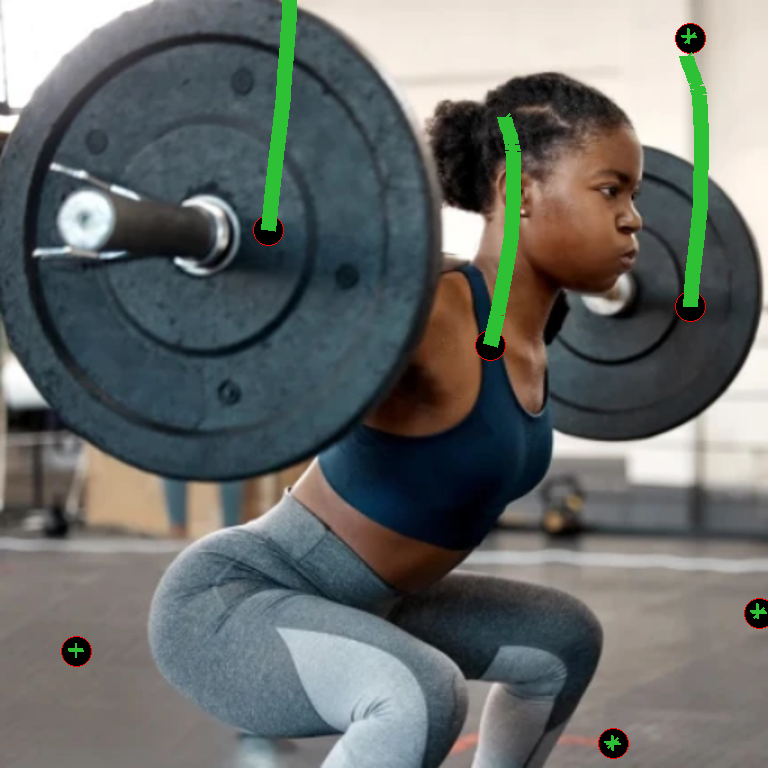}{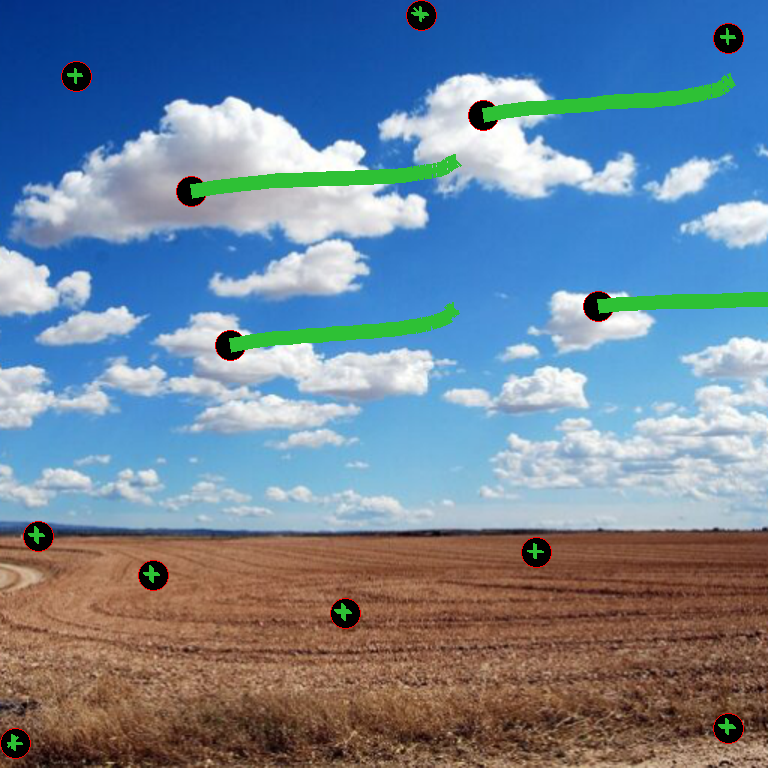}\\
        \imagerow{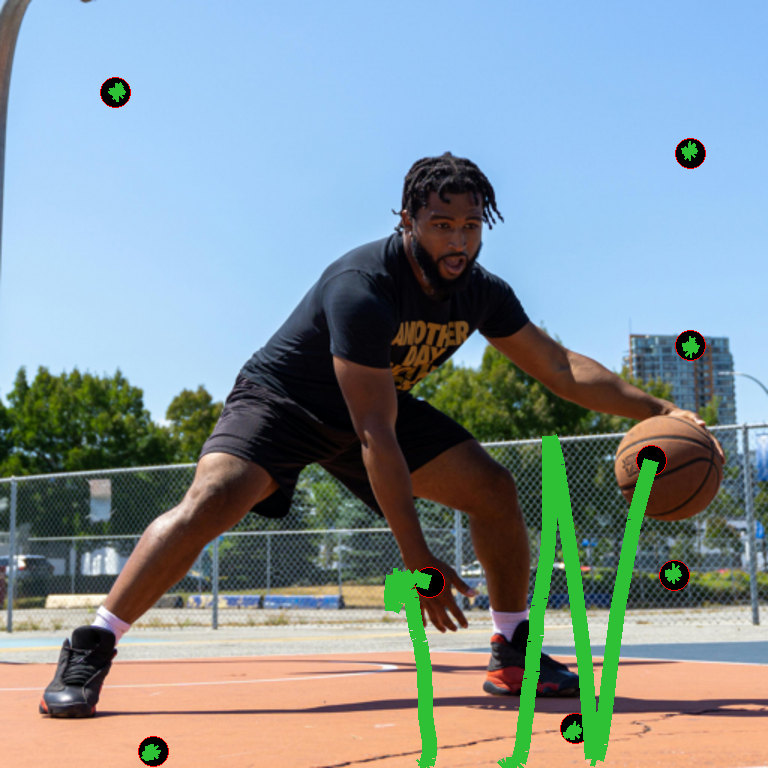}{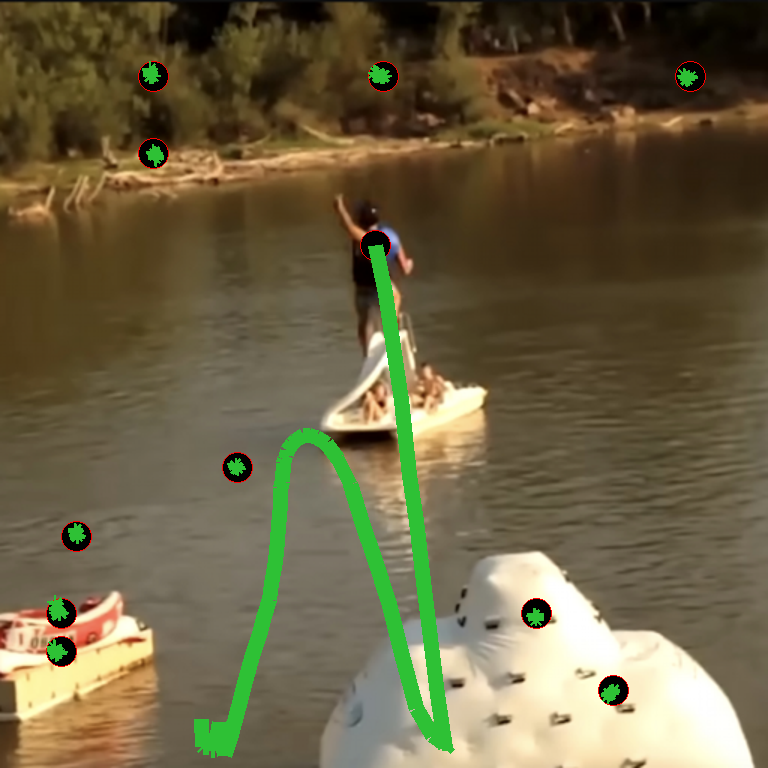}{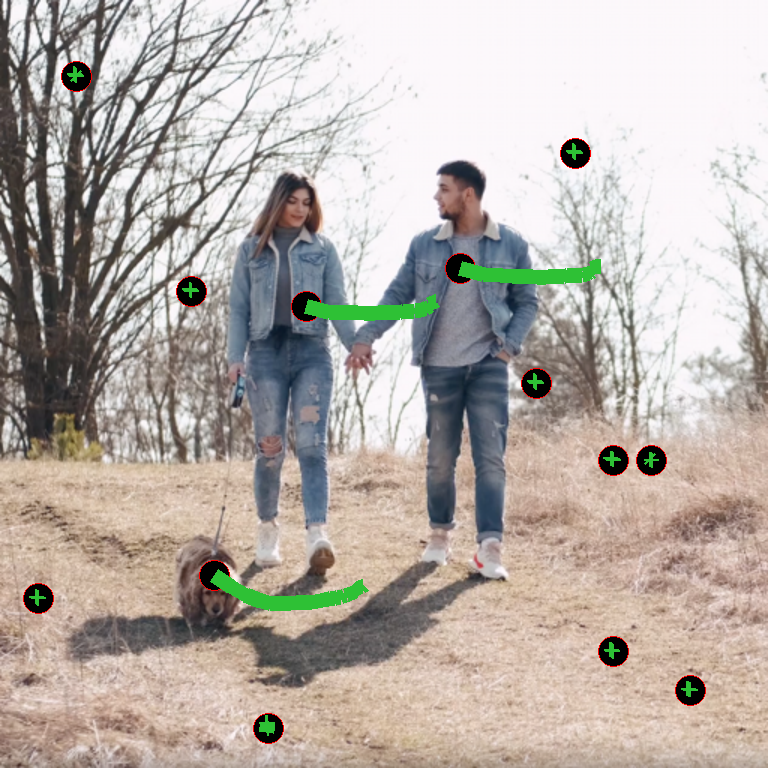}{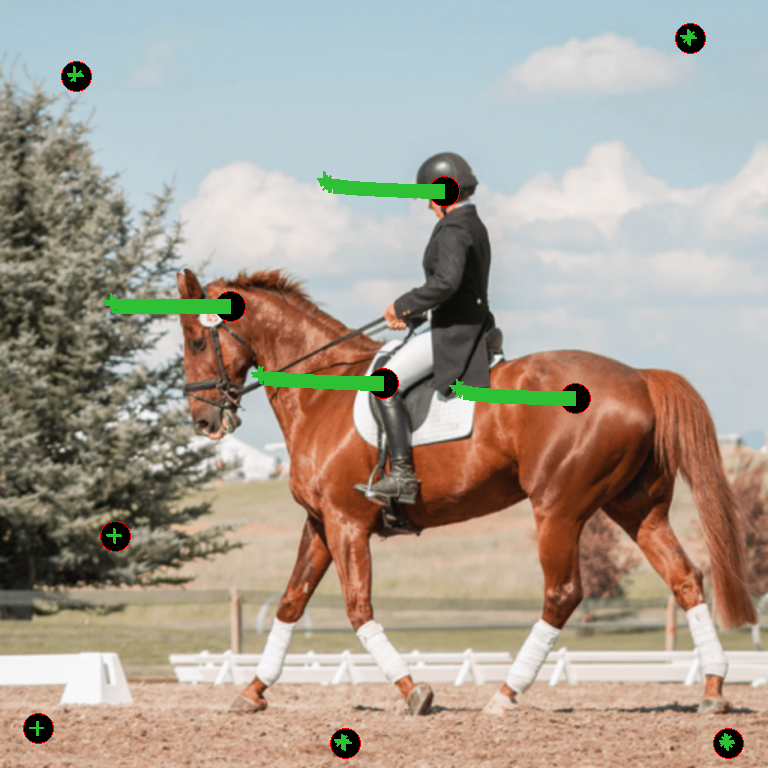}\\
    \end{tabular}
    \caption{\textbf{Hinted Samples} Our model is capable of producing complex, coherent, and appearance-informed motion given an initial motion hint.}
    \label{fig:poked_qual}
\end{figure*}

\cref{fig:poked_qual} shows samples generated with initial motion hints. The samples show that unrolling hinted trajectories results in consistent motion across entities and realistic long-term behaviour.

\begin{figure*}[t]
    \newcommand{\imgwidth}{0.175\linewidth}
    \newcommand{\insertimage}[1]{\includegraphics[width=\imgwidth]{fig/qual/unpoked/#1_unpoked.jpg}}
    \newcommand{\imagerow}[4]{\insertimage{#1}&\insertimage{#2}&\insertimage{#3}&\insertimage{#4}}
    \centering
    \begin{tabular}{cccc}
        \imagerow{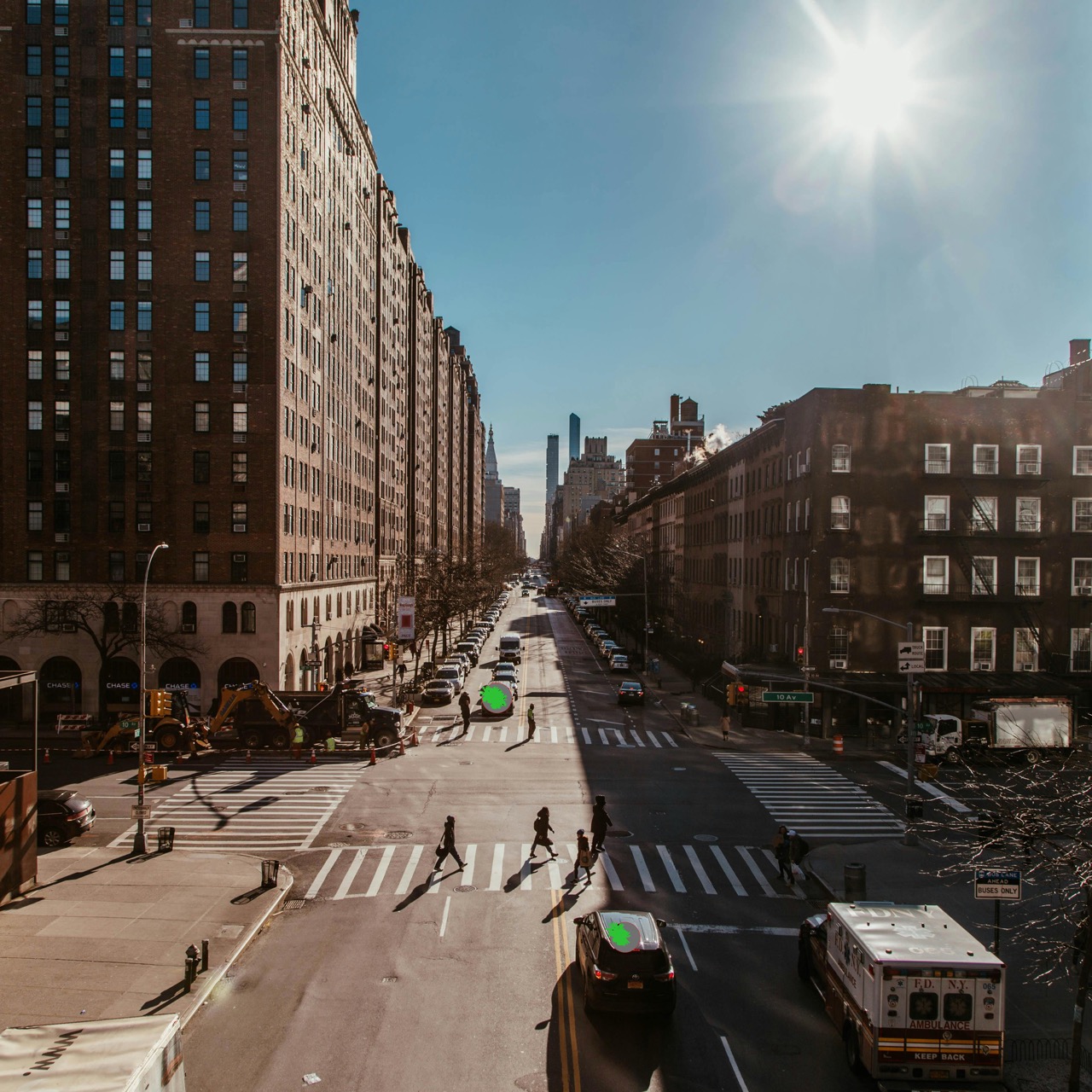}{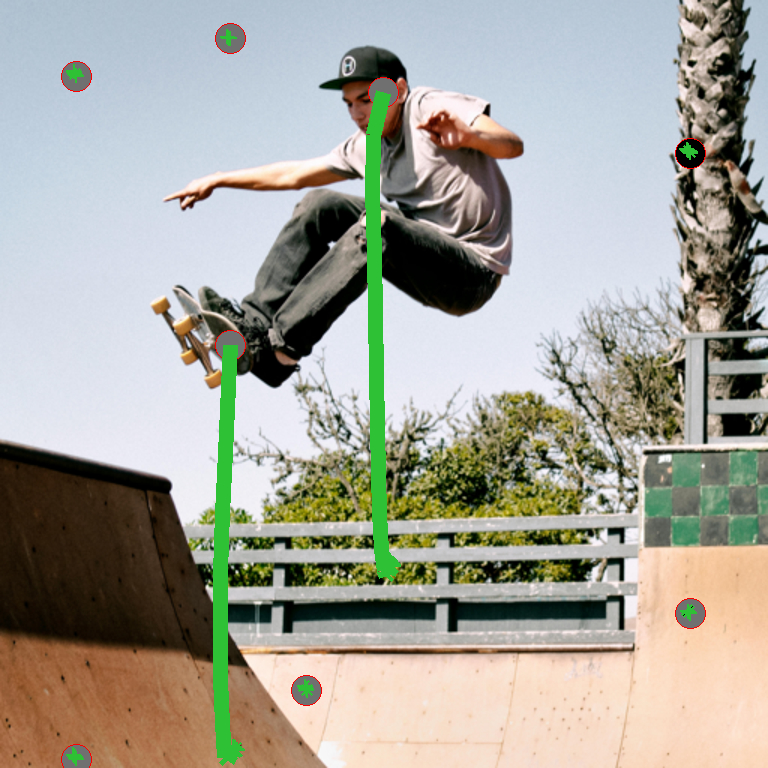}{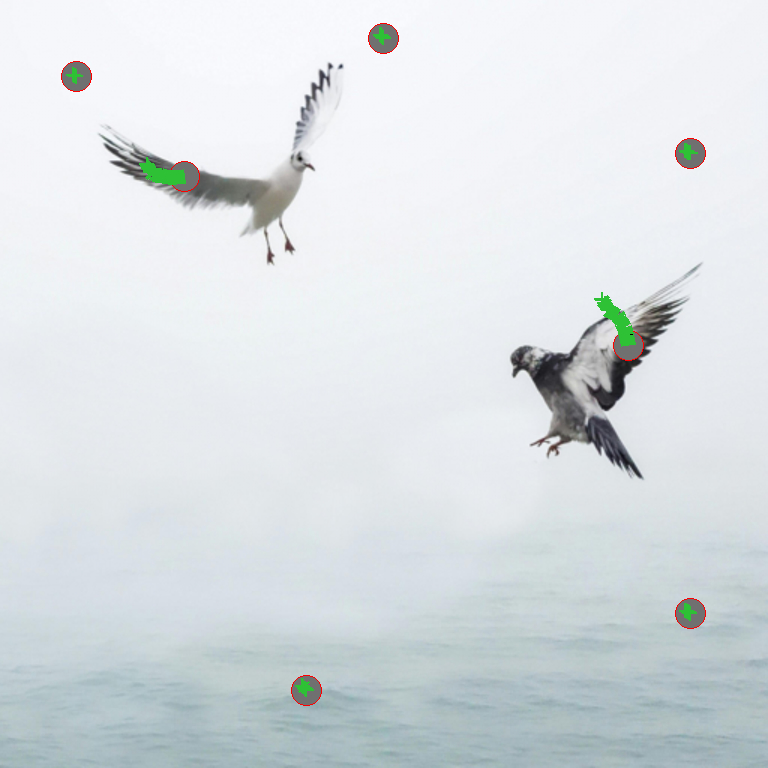}{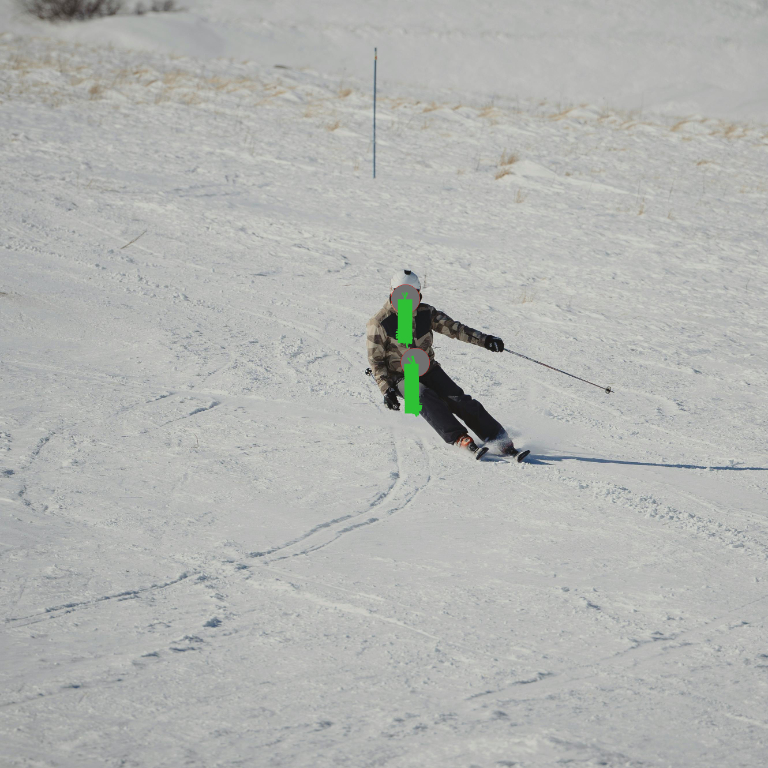}\\
    \end{tabular}
    \caption{\textbf{Un-hinted Samples} Given only appearance conditioning, our approach is able to produce physically correct and coherent motion, also show ing more complex understanding, such as that cars at an intersection should \textit{not} move when pedestrians are blocking their path.}
    \label{fig:unpoked_qual}
\end{figure*}

In \cref{fig:unpoked_qual} we provide samples generated without an initial motion hint. While the trajectories tend to be more simplistic, realistic motion is obtained based solely on the input image. Query points without motion hint are marked in grey.

\begin{figure*}[t]
    \newcommand{\imgwidth}{0.175\linewidth}
    \newcommand{\insertimage}[1]{\includegraphics[width=\imgwidth]{fig/qual/partially_poked/#1_partially_poked.jpg}}
    \newcommand{\imagerow}[4]{\insertimage{#1}&\insertimage{#2}&\insertimage{#3}&\insertimage{#4}}
    \centering
    \begin{tabular}{cccc}
        \imagerow{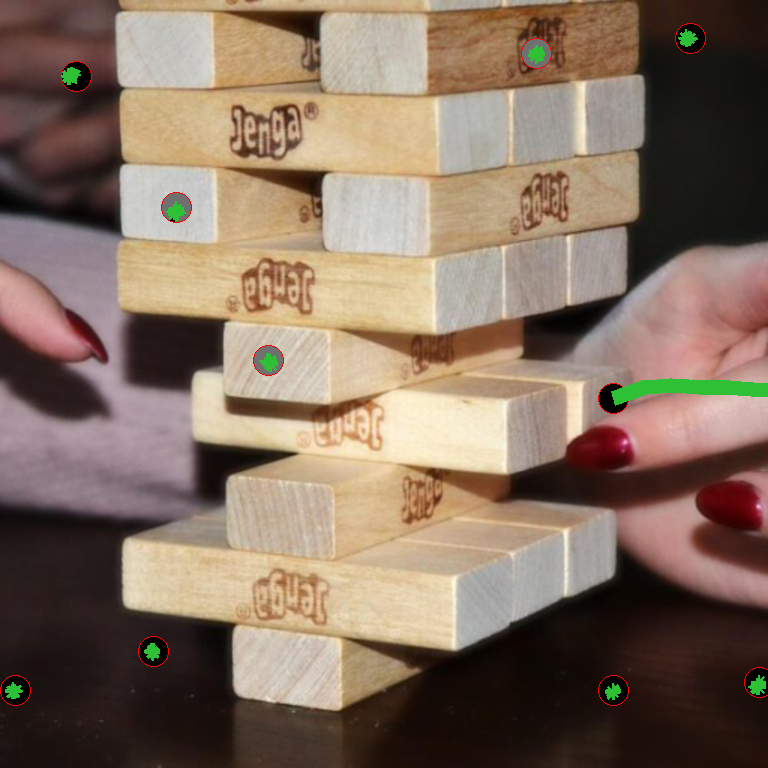}{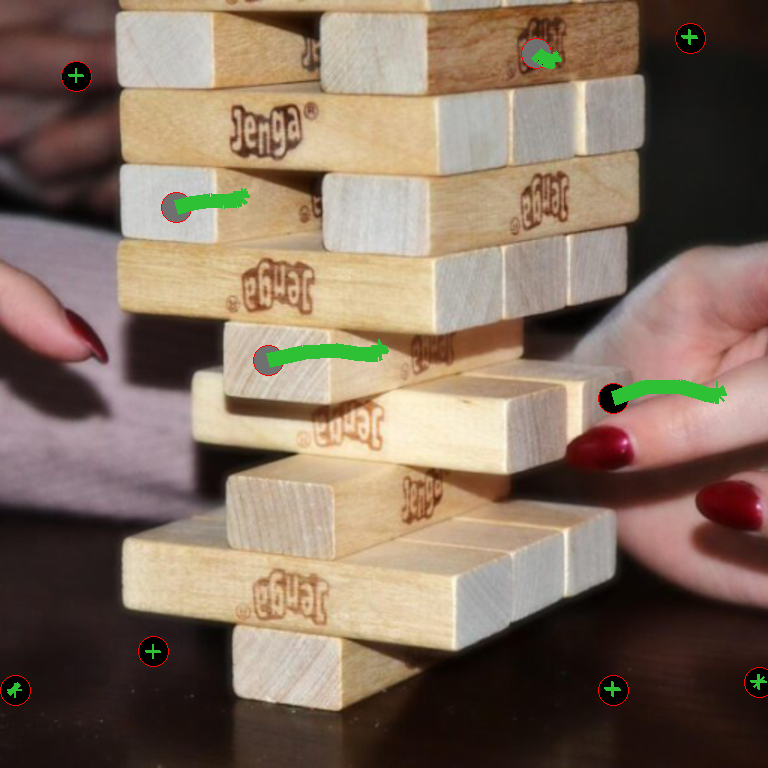}{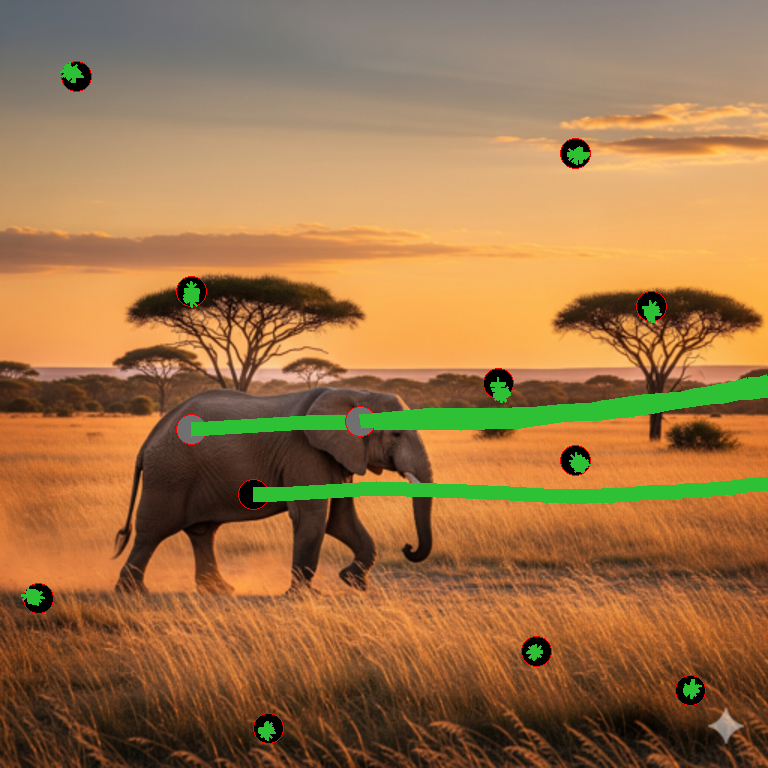}{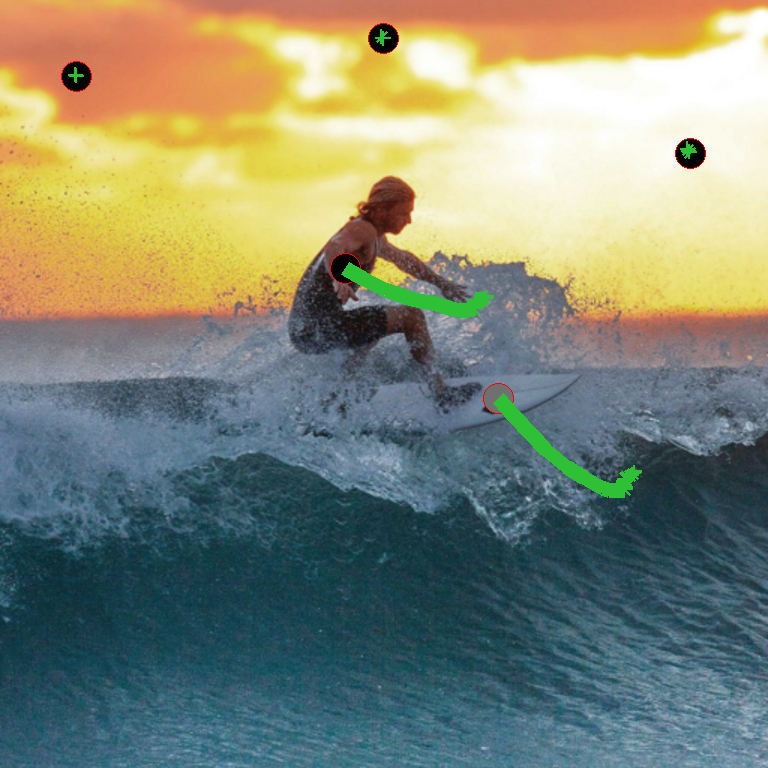}\\
    \end{tabular}
    \caption{\textbf{Partially-hinted Samples} Given only a single poke conditioning per example, our model produces coherent motion for queries on the same or linked objects. {The Jenga example highlights that our model is able to capture multi-modality if two outcomes are possible given the same initial motion hint.}}
    \label{fig:partially_poked_qual}
\end{figure*}

\cref{fig:partially_poked_qual} illustrates samples where only a single query point on the object received a hint, and motion for other queries has to be inferred from appearance alone. The results highlight that sampled motion is coherent across objects. Further, our model is able to capture multi-modal behaviour if two outcomes are realistic given the same input. Query points without motion hint are marked in grey while queries with motion hint are colored black.

\billiardsamples{
    \paragraph{Billiard samples}
    We show qualitative predictions from our model trained on billiard simulation data in \cref{fig:comparison_billiard}. We show the predicted simulation given an initial impulse in the upper row, and the ground truth simulation overlaid with the prediction in the lower row for each respective sample. Simulation time increases linearly from left to right. Our model is able to accurately predict the ground truth motion, up to stochastic uncertainties.
}

\section{Language Model Usage}
We employed large language models (OpenAI GPT-5.2, Claude Opus 4.6) for text refinement purposes, including improving grammar and as inspiration for rephrasing sections.
They were also employed to provide feedback on early drafts and propose initial implementations for auxiliary utility functions not directly related to the paper's contributions, subsequently verified and reworked by the authors.
No scientific content, experimental results, or novel ideas were generated by LLMs -- all technical contributions were conceived, implemented, and verified by the authors.

\clearpage

\billiardsamples{
    \begin{figure*}[h]
    \centering
    \includegraphics[width=\textwidth]{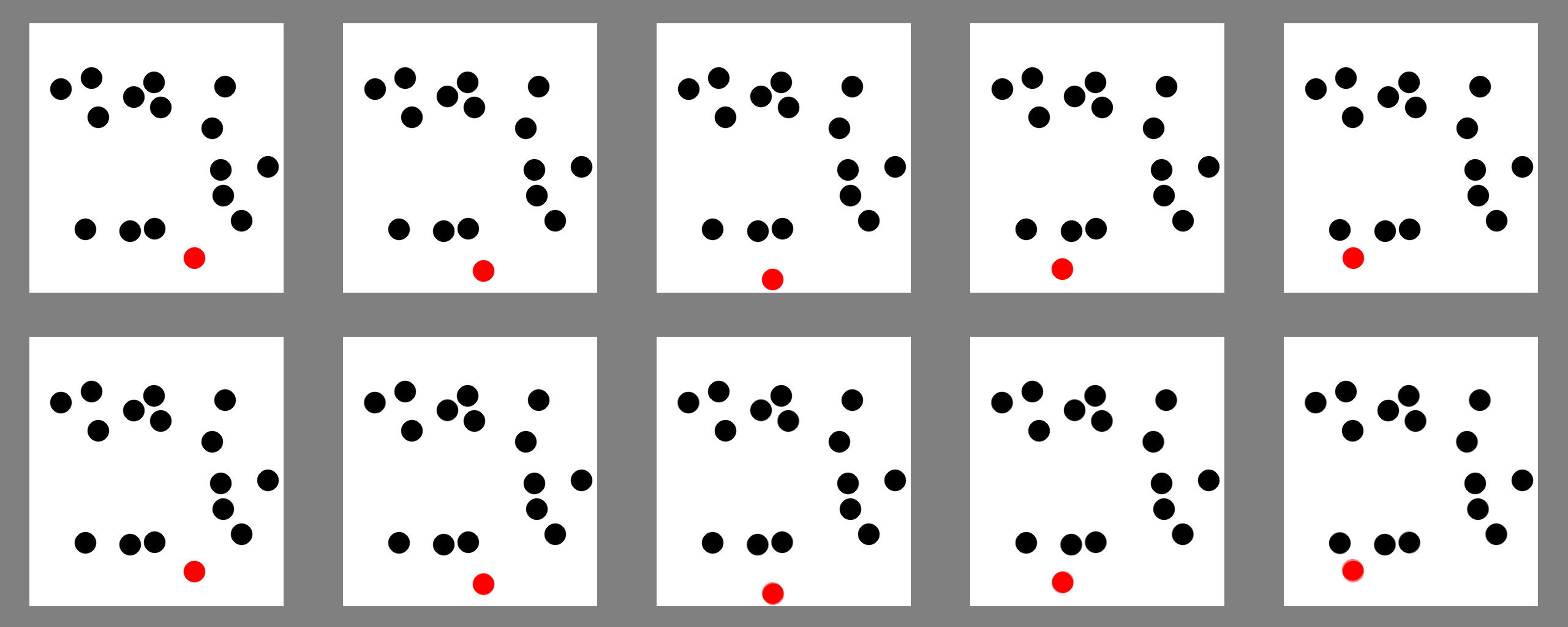}
    \includegraphics[width=\textwidth]{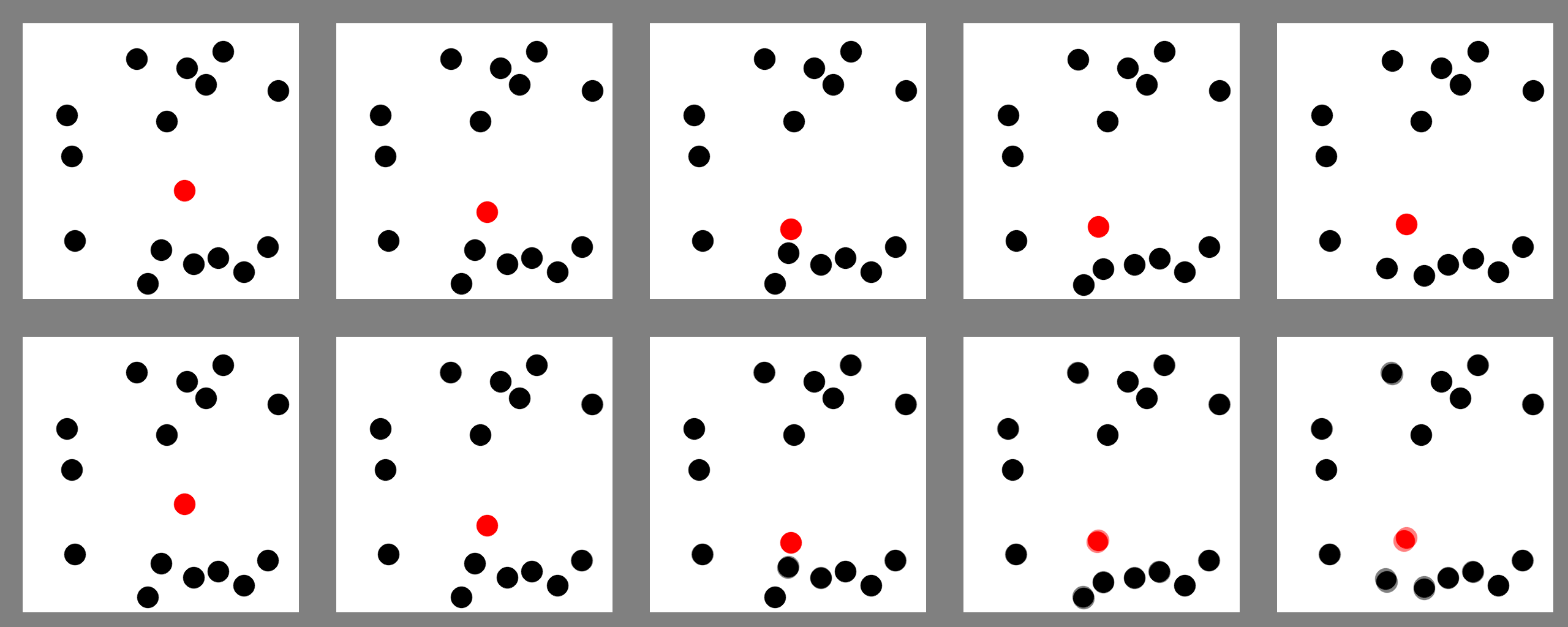}
    \includegraphics[width=\textwidth]{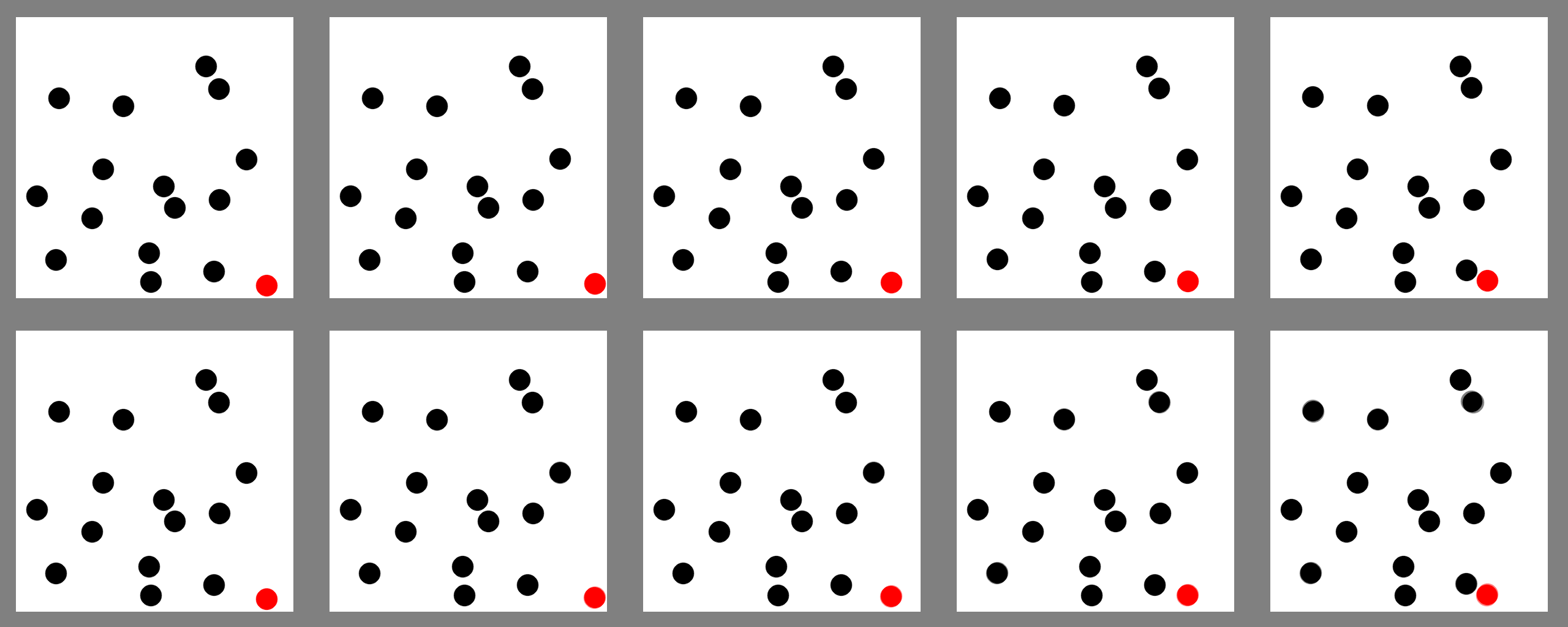}
    \caption{\textbf{Qualitative samples on our billiard simulation.} The respective top row shows our model's prediction given an initial impulse for the ball marked in red, where we visualize the predicted trajectory state using a frame-wise renderer; the lower row shows an overlay of the ground truth simulation with the prediction to enable comparisons. Our model can successfully predict the observed motion up to minor stochastic details.}
    \label{fig:comparison_billiard}
    \end{figure*}
}

\clearpage

\end{document}